\documentclass[twoside]{article}

\usepackage[accepted]{aistats2024}
\usepackage[round]{natbib}
\renewcommand{\bibname}{References}
\renewcommand{\bibsection}{\subsubsection*{\bibname}}

\usepackage{amsmath,amsfonts,bm}

\def\eqref#1{Eq.~\ref{#1}}

\def\1{\bm{1}}

\def\rva{{\mathbf{a}}}
\def\rvb{{\mathbf{b}}}

\def\rve{{\mathbf{e}}}

\def\rvg{{\mathbf{g}}}
\def\rvh{{\mathbf{h}}}
\def\rvu{{\mathbf{i}}}

\def\rvq{{\mathbf{q}}}

\def\rvs{{\mathbf{s}}}

\def\rvu{{\mathbf{u}}}

\def\rvx{{\mathbf{x}}}
\def\rvy{{\mathbf{y}}}
\def\rvz{{\mathbf{z}}}

\def\rmA{{\mathbf{A}}}

\def\rmC{{\mathbf{C}}}
\def\rmD{{\mathbf{D}}}
\def\rmE{{\mathbf{E}}}

\def\rmH{{\mathbf{H}}}
\def\rmI{{\mathbf{I}}}

\def\rmL{{\mathbf{L}}}

\def\rmS{{\mathbf{S}}}

\def\rmW{{\mathbf{W}}}
\def\rmX{{\mathbf{X}}}
\def\rmY{{\mathbf{Y}}}

\def\vmu{{\bm{\mu}}}

\def\va{{\bm{a}}}

\DeclareMathAlphabet{\mathsfit}{\encodingdefault}{\sfdefault}{m}{sl}
\SetMathAlphabet{\mathsfit}{bold}{\encodingdefault}{\sfdefault}{bx}{n}

\def\gA{{\mathcal{A}}}

\def\gD{{\mathcal{D}}}
\def\gE{{\mathcal{E}}}

\def\gG{{\mathcal{G}}}
\def\gH{{\mathcal{H}}}
\def\gI{{\mathcal{I}}}

\def\gL{{\mathcal{L}}}

\def\gN{{\mathcal{N}}}

\def\gR{{\mathcal{R}}}

\def\gV{{\mathcal{V}}}

\def\sR{{\mathbb{R}}}

\newcommand{\R}{\mathbb{R}}

\usepackage{hyperref}
\usepackage{url}

\usepackage[utf8]{inputenc} 
\usepackage[T1]{fontenc}    
\usepackage{hyperref}       
\usepackage{url}            
\usepackage{booktabs}       
\usepackage{amsfonts}       
\usepackage{nicefrac}       
\usepackage{microtype}      
\usepackage{xcolor}         
\usepackage{amsmath}
\usepackage{float}
\usepackage{multirow}
\usepackage{graphicx} 
\usepackage{wrapfig}
\usepackage{xspace}
\usepackage{caption}
\usepackage{afterpage}

\definecolor{ballblue}{rgb}{0.13, 0.67, 0.8}

\newcommand{\model}{BloomGML\xspace}

\newtheorem{definition}{Definition}

\newtheorem{proposition}{Proposition}

\newcommand{\h}[1]{\rvh^{(#1)}}

\newcommand{\elow}{\ell_{low}}

\input{def.set}

\begin{document}

%

%
\runningauthor{Zheng, He, Qiu, Wang, Wipf}

\twocolumn[

\aistatstitle{Graph Machine Learning through the Lens of Bilevel Optimization}

\vspace*{-0.3cm}

\aistatsauthor{ Amber Yijia Zheng\footnotemark \\Purdue University \And Tong He \\ Amazon Web Services \And  Yixuan Qiu\\ Shanghai University of\\Finance and Economics \AND Minjie Wang \\ Amazon Web Services \And David Wipf \\ Amazon Web Services}

\aistatsaddress{}]

\footnotetext{Contribution during AWS Shanghai AI  Lab internship.}

\vspace*{-0.7cm}
\begin{abstract}
\vspace*{-0.4cm}
Bilevel optimization refers to scenarios whereby the optimal solution of a lower-level energy function serves as input features to an upper-level objective of interest.  These optimal features typically depend on tunable parameters of the lower-level energy in such a way that the entire bilevel pipeline can be trained end-to-end.  Although not generally presented as such, this paper demonstrates how a variety of graph learning techniques can be recast as special cases of bilevel optimization or simplifications thereof.  In brief, building on prior work we first derive a more flexible class of energy functions that, when paired with various descent steps (e.g., gradient descent, proximal methods, momentum, etc.), form graph neural network (GNN) message-passing layers; critically, we also carefully unpack where any residual approximation error lies with respect to the underlying constituent message-passing functions.  We then probe several simplifications of this framework to derive close connections with non-GNN-based graph learning approaches, including knowledge graph embeddings, various forms of label propagation, and efficient graph-regularized MLP models. And finally, we present supporting empirical results that demonstrate the versatility of the proposed bilevel lens, which we refer to as \textit{\model}, referencing that \textit{BiLevel Optimization Offers More Graph Machine Learning}. Our code is available at \url{https://github.com/amberyzheng/BloomGML}.  Let graph ML \textit{bloom}.
\end{abstract}

\vspace{-0.5cm}
\section{Introduction}
\vspace{-0.2cm}

Graph machine learning covers a wide range of modeling tasks involving graph-structured data, where cross-instance/node dependencies are reflected by edges.  As a classical example, label propagation \citep{zhou2003learning} and its many offshoots represent a semi-supervised learning approach whereby observed node labels are iteratively spread across the graph to unlabeled nodes.  In a related vein, various forms of graph-regularized MLP models \citep{ando2006learning,hu2021graph,zhang2023orthoreg} share node representations across edges to penalize misalignment with network structure during training.  And as a third example more narrowly focused on certain heterogeneous graphs, non-parametric knowledge graph embedding (KGE) models \citep{bordes2013translating} produce node and relation-type embeddings that have been trained to differentiate factual knowledge triplets, composed of head and tail nodes connected by an edge relation, from spurious ones.

More recently, graph neural networks (GNNs) have emerged as a promising class of predictive models for handling tasks such as node classification or link prediction \citep{kipf2016semi,hamilton2017inductive,xu2019powerful,velivckovic2017graph,zhou2020graph}.  Central to a wide variety of GNN architectures are layers composed of three components: a \emph{message function}, which bundles information for sharing with neighbors, an \emph{aggregation function} that fuses all the messages from neighbors, and an \emph{update function} that computes the layer-wise output embedding for each node.  Collectively, these functions enable the layer-by-layer propagation of information across the graph to facilitate downstream tasks \citep{kipf2016semi,hamilton2017inductive,kearnes2016molecular}.

In the past, the graph ML models described above have primarily been motivated from diverse perspectives, without necessarily a single, transparent lens with which to evaluate their commonalities.  To make strides towards a more cohesive graph ML narrative, we propose to operate from a unifying vantage point afforded by bilevel optimization \citep{7942105,wang2016learning}.  The latter refers to an optimization problem characterized by two interconnected levels, whereby the optimal solution of a \emph{lower-level} objective serves as input features to an \emph{upper-level} loss.  Within the context of graph ML, we will demonstrate how the lower-level in particular can induce interpretable model architectures, while exposing underappreciated similarities across seemingly disparate paradigms.  After providing relevant background material in Section \ref{sec:background}, our primary contributions throughout the remainder of the paper can be summarized as follows:
\begin{itemize}

\item In Section \ref{sec:gen_bilevel_gnn} we introduce a broad class of lower-level energy functions that, when combined with appropriate optimization steps, give rise to efficient GNN message-passing layers inheriting interpretable inductive biases of energy minimizers.  Moreover, subject to certain technical assumptions, we demonstrate that these optimization- induced layers can replicate arbitrary message and aggregation functions, while providing a flexible approximation for the update function. Hence we conditionally isolate any appreciable approximation error to the latter, and in doing so, articulate a transparent design space for GNN architectures produced by bilevel optimization.


\item Based on the above, in Section \ref{sec:broader_graph_ML} we propose a broadly-applicable bilevel optimization framework called \model, and establish that several notable special cases effectively reproduce and unify non-GNN-based graph learning approaches.  In particular, we demonstrate an equivalence between traditional KGE learning and the training of an implicit GNN, with parameters given by relational embeddings and a graph formed from positive and negative knowledge triplets.  This association explains recent empirical findings in the literature while motivating new interpretable approaches to modeling over knowledge graphs.

\item Finally, in Section \ref{sec:experiments} we present experiments that highlight the flexibility and explanatory power of candidate models motivated by our proposed bilevel optimization framework and attendant analysis thereof.
\end{itemize}

\vspace*{-0.4cm}
\section{Background} \label{sec:background}
\vspace*{-0.2cm}
In this section, we first introduce GNN architectures and their constituent message-passing layers as applied to node classification tasks (more general tasks will be addressed later).  We then describe an existing, popular special case induced by bilevel optimization, followed by a discussion of both advantages and limitations.

\vspace*{-0.1cm}
\subsection{Message Passing Graph Neural Networks}
\vspace*{-0.1cm}
Let $\gG = \{\gV,\gR,\gE \}$ denote a heterogeneous graph with $n = |\gV|$ nodes and $m = |\gR|$ edge types.  Each edge $e \in \gE$ is composed of a triplet via $ e = (u, r, v)$, with nodes $u,v \in \gV$ and relation type $r \in \gR$.  The more general heterogeneous case defaults to a homogeneous graph when $m = 1$.  For any given node $v$, we also define the set of (1-hop) neighbors as $\mathcal{N}_v := \{(u, r): (u, r, v) \in \gE \}$.  Additionally, associated with each node is a $d$-dimensional feature vector and a $c$-dimensional label vector which stack to form $\rmX \in \mathbb{R}^{n\times d}$ and $\rmY \in\R^{n\times c}$, respectively.  The latter reflects our focus on node classification tasks within this section for simplicity of exposition; the extension to link prediction (whereby labels correspond with supervision over edges rather than individual nodes) follows naturally. 

When presented with a graph $\gG$ so defined, a message-passing GNN (or MP-GNN) of depth $L$ produces a layer-wise sequence of node embeddings $\{\rmH^{(l)} \}_{l=1}^L$, where $\rmH^{(l)} = \{\rvh_v^{(l)} \}_{v \in \gV} \in \mathbb{R}^{n \times d}$ represents the aggregated set of embeddings $\rvh_v^{(l)}$ at the $l$-th layer.  Moreover, for each $l$, $\rmH^{(l)}$ is computed from $\rmH^{(l-1)}$ using a composition of three functions characteristic of MP-GNN architectures \citep{messagepassing,hamilton2020graph}.  We formalize this composition via the following definition:

\begin{definition} \label{def:mpgnn_layer}
    An \textit{MP-GNN layer} is formed via the composition of three functions:
    \vspace*{-0.2cm}
    \begin{enumerate}
        \item A \underline{message function} $f_M$ that computes $\vmu^{(l)}_{(u, r, v)}:= f_M(\h{l-1}_u, r, \h{l-1}_v) \in \mathbb{R}^d$ for any edge $e = (u, r, v) \in \gE$, i.e., given the head and tail nodes and the relation type, $f_M$ produces an embedding for the corresponding edge;
        \item A permutation-invariant \underline{aggregation function} $f_A$ that computes $\rva^{(l)}_v := f_A(\{\vmu^{(l)}_{(u, r, v)}: (u, r) \in \mathcal{N}_v\}) \in \mathbb{R}^d$ for all $v \in \gV$, meaning for each node the set function $f_A$ aggregates all messages associated with other nodes sharing an edge;
        \item An \underline{update function} $f_U$ that computes the embeddings of the next layer as $\rvh^{(l)}_v = f_U(\rvh^{(l-1)}_v, \rva^{(l)}_v, \rvx_v)$ for all $v \in \gV$, i.e., the embeddings from the previous layer are combined with the aggregated messages and (optionally) the original input node features $\rvx_v \in \rmX$.
    \end{enumerate}
\end{definition}
\vspace*{-0.2cm}
The permutation-invariance of $f_A$ notwithstanding, all three components defining the MP-GNN layer can in principle be any parameterized differentiable (almost everywhere) functions suitable for learning with SGD, etc.  Hence different MP-GNN architectures are more-or-less tantamount to different selections for $f_M$, $f_A$, and $f_U$.  We adopt $\rmW$ to denote the bundled set of all trainable parameters within $\{f_M, f_A, f_U \}$.





Conditioned on executing $L$ layers per the schema of Definition \ref{def:mpgnn_layer} culminating in the embeddings $\rmH^{(L)}$, model training proceeds by minimizing the loss
\begin{equation}
\label{eq:general_bilevel_upper}
\ell_{up}(\rmW,\Theta) := \sum_{v \in \gV'} \gD\left[ g \left(\rvh^{(L)}_v(\rmW) ; \Theta \right), \rvy_v\right],
\end{equation}
where $\rvh^{(L)}_v(\rmW) \equiv \rvh^{(L)}_v$ reflects the explicit dependency of $\rmH^{(L)}$ on $\rmW$.  In this expression,  the output layer $g: \mathbb{R}^d \rightarrow \mathbb{R}^c$ is a differentiable node-wise function parameterized by $\Theta$, $\gV' \subset \gV$ indicates the subset of training instances with observable labels, and $\mathcal{D}$ is a discriminator function, e.g., cross-entropy for classification.  Finally, the ``up'' in $\ell_{up}$ reflects the fact that this function will later serve as an \textit{upper-level} loss when embedded within a bilevel optimization setting to be described next.


\vspace*{-0.2cm}
\subsection{Graph-Centric Bilevel Optimization} \label{sec:bilevel_example}
\vspace*{-0.2cm}
Our generic MP-GNN setting thus far narrows to the realm of bilevel optimization \citep{7942105} when we infuse the latent embeddings $\{ \rmH^{(l)} \}_{l=1}^L$ from above with additional layer-wise structure determined by a second, \textit{lower-level} loss $\ell_{low}(~\cdot~; \rmW,\gG)$, which likewise depends on $\rmW$ and $\gG$. Specifically, suppose that for all layers $l$, we enforce that
\begin{equation} \label{eq:low_level_criteria}
\begin{split}
    & \ell_{low}\left( \rmH^{(l)}; \rmW, \gG \right)    \leq  \ell_{low}\left( \rmH^{(l-1)}; \rmW, \gG  \right), \\
    & ~\mbox{and ideally}~~\rmH^{(L)} \approx  \underset{\rmH}{\arg\min} \ \ell_{low}\left( \rmH; \rmW, \gG \right)
\end{split} 
\end{equation}
for $L$ sufficiently large.  In other words, at each layer the composite embedding model $f_U \circ f_A \circ f_M$ that updates $\rmH^{(l)}$ per Definition \ref{def:mpgnn_layer} dually serves as an algorithm producing descent steps along the loss surface of $\ell_{low}$; hence the coincident dependency of both $\ell_{low}$ and $f_U \circ f_A \circ f_M$ on $\rmW$ and $\gG$.  When combined with (\ref{eq:general_bilevel_upper}), we arrive at our bilevel destination, with embeddings that iteratively minimize $\ell_{low}$ forming predictive features for end-to-end training within $\ell_{up}$.

\paragraph{\textbf{A Representative Special Case.}}  Motivated by \cite{zhou2003learning} in the context of homogeneous graphs, we consider the energy $\ell_{low}(\rmH;\rmW,\gG) := $
\begin{equation} \label{eq:quad_example_energy}
\sum_{v \in \gV} \tfrac{1}{2}\left\|\rvh_v - \pi(\rvx_v; \rmW)  \right\|_{2}^2 + \tfrac{\lambda}{2} \sum_{(u,v) \in \gE} \left\|\rvh_u - \rvh_v \right\|_{2}^2,
\end{equation}
where $\pi$ is a parameterized input model (e.g., an MLP with weights $\rmW$), and $\lambda > 0$ is a trade-off parameter.  This factor balances local consistency w.r.t.~the input model $\pi$  and global smoothness across the graph.

Next, starting from some $\rmH^{(0)}$, we may iterate $L$ gradient descent steps along (\ref{eq:quad_example_energy}) to obtain $\rmH^{(L)}$.  And along this trajectory for each layer $l$, it is straightforward to show (see Appendix \ref{sec:derive_mp_special_case}) that the resultant updates for each node $v$ adhere to the stipulations of Definition \ref{def:mpgnn_layer}, and can be equivalently computed as 
\begin{eqnarray} \label{eq:simple_mp_case}
    && \hspace*{-0.7cm} \vmu^{(l)}_{(u,r, v)} = \rvh_u^{(l-1)} - \rvh_v^{(l-1)}, ~~ \rva^{(l)}_v = \sum_{u \in \mathcal{N}_v} \vmu^{(l)}_{(u,r, v)}, ~~\mbox{and} \nonumber \\ 
    && \hspace*{-0.7cm} \rvh_v^{(l)} = (1-\gamma) \rvh_v^{(l-1)}  + \gamma \lambda  \rva^{(l)}_v + \gamma \pi(\rvx_v;\rmW),
\end{eqnarray}
where $\gamma$ is the gradient step-size parameter. From this expression, we observe that the effective message-passing layer involves weighted skip connections from the previous layer and input model, along with a permutation-invariant summation of the embeddings from neighboring nodes.  And provided $\pi$ is differentiable w.r.t.~$\rmW$, the update for $\rvh_v^{(l)}$ computed via (\ref{eq:simple_mp_case}) is differentiable as well.  Hence we can substitute into (\ref{eq:general_bilevel_upper}) and train the entire system end-to-end using SGD over $\rmW$ and $\Theta$.


Using various preconditioners and reparameterizations, it has been widely noted that model layers constructed via (\ref{eq:simple_mp_case}), or variations thereof, directly correspond with various popular MP-GNN architectures \citep{chen2021does,ma2021unified,pan2020unified,xue2023lazygnn,yang2021graph,zhang2020revisiting,zhu2021interpreting,zhou2021dirichlet,di2023understanding}.  Hence this particular instance of bilevel optimization can be interpreted as training a GNN node classification model with layers designed to minimize (\ref{eq:quad_example_energy}).

\vspace*{-0.2cm}
\paragraph{\textbf{Why Construct GNN Architectures this Way?}}





As one notable example, across customer-facing applications it is often desirable to know which factors influence the output predictions of a GNN model  \citep{ying2019gnnexplainer}.  Relevant to providing such explanatory details, the embeddings produced by bilevel optimization as in Section \ref{sec:bilevel_example} contain additional information when contextualized w.r.t.~their role in the $\ell_{low}$.  For instance, if some node embedding $\rvh_v^{(L)}$ within (\ref{eq:quad_example_energy}) is very far from $\pi(\rvx_v;\rmW)$ relative to other nodes, it increases our confidence that subsequent predictions may be based on network effects rather than an irrelevant local feature $\rvx_v$.  We will empirically explore this possibility in Section \ref{sec:experiments}.  Alternatively, in the context of heterophilic graphs, we can examine the distribution of $\|\rvh_v - \rvh_u \|_2$ across nodes sharing an edge to loosely calibrate the degree of heterophily, and possibly counteract its impact via 
appropriate modifications.  And as a final dimension of motivation, the bilevel perspective provides natural entry points for the scalable training of models with arbitrary depth \citep{xue2023lazygnn} or interpretable integration with offline sampling \citep{jiang2023musegnn}.


\vspace*{-0.2cm}
\subsection{Unresolved Limitations}
\vspace*{-0.2cm}
Despite the widespread use of bilevel optimization in forming and/or analyzing GNN layers, prior work is largely confined to the exploration of narrow instances that follow from quite specific (usually quadratic) choices for $\ell_{low}$ paired with vanilla gradient descent, with application to homogeneous graphs; the example presented in Section \ref{sec:bilevel_example} follows this paradigm.  As of yet, there has been no clear delineation of a broad design space of possibilities.  We take initial steps in this direction next.

\vspace*{-0.2cm}
\section{Towards More Flexible GNNs from Bilevel Optimization} \label{sec:gen_bilevel_gnn}
\vspace*{-0.2cm}





In the previous section, we applied a \textit{specific} iterative algorithm, gradient descent with step-size $\gamma$, to a \textit{particular} energy function, $\ell_{low}(\rmH;\rmW,\gG)$ from (\ref{eq:quad_example_energy}), and the resulting descent updates produced the GNN message-passing layers given by (\ref{eq:simple_mp_case}).  Within this relatively narrow design space, the only algorithmic flexibility is in the choice of $\gamma$, and on the energy side, we are limited to tuning the trade-off parameter $\lambda$ and the input model $\pi$.  Not surprisingly then, (\ref{eq:simple_mp_case}) is far less flexible/expressive relative to the general form from Definition \ref{def:mpgnn_layer}.  We now seek to reduce this gap via the following formulation.

To begin we require some additional notation.  Let $\gL(\gG)$ be a function space of interest dependent on graph $\gG$, where each function $\ell_{low} \in \gL(\gG)$ is specified as $\ell_{low} : \gH \rightarrow \mathbb{R}$ over some node embedding domain $\gH \subseteq \mathbb{R}^{n\times d}$.  We then define  $\gA : \gL(\gG) \times \gH \rightarrow \gH$ as an arbitrary iterative mapping/algorithm that satisfies
\begin{equation} \label{eq:descent_algo_abstract}
    \ell_{low}\left(~\gA[\ell_{low}, \rmH] ~\right) ~ \leq ~ \ell_{low}(\rmH) ~~~ \forall \ell_{low} \in \gL(\gG),~ \forall \rmH \in \gH.
\end{equation}
This expression merely entails that if we apply $\gA$ to $\ell_{low}$ initialized at $\rmH$, the resulting iteration produces embeddings that reduce (or leave unchanged) $\ell_{low}$.  We are then poised to ask a central motivating question:

\textit{Can we find a sufficiently flexible pairing of some $\calA$ and compatible $\ell_{low} \in \calL(\calG)$ such that $\gA[\ell_{low}, \rmH]$ closely approximates, to the extent possible, any composite message-passing layer $f_U \circ f_A \circ f_M$ which follows from the stipulations of Definition \ref{def:mpgnn_layer}?}  

To make initial progress in answering this question, we first convert Definition \ref{def:mpgnn_layer} to what we will refer to as a \textit{canonical form} designed to minimize, without loss of generality, the non-uniqueness of the decomposition of $f_U \circ f_A \circ f_M$ into respective message, aggregation, and update functions.  In so doing we obtain a more transparent entry point for isolating precisely where more-or-less exact alignment between $\gA[\ell, \rmH]$ and $f_U \circ f_A \circ f_M$ is possible, and where non-trivial gaps remain.

\subsection{Canonical Form of MP-GNN Layers}

Given the composite function $f_U \circ f_A \circ f_M$, there is no unique decomposition into its constituent parts, even within the constraints of Definition \ref{def:mpgnn_layer}.  To this end, we introduce the following simplification of MP-GNN layers that, as we will later show, leads to a unique decomposition (up to inconsequential linear transformations) while maintaining expressiveness:
\begin{definition} \label{def:mpgnn_layer_canon}
    An MP-GNN layer in \textit{canonical form} is defined as the composition of three functions:
    \begin{enumerate}
        \item A \underline{message function} $\widetilde{f}_M$ that computes $\vmu^{(l)}_{(u, r, v)}:= \widetilde{f}_M(\h{l-1}_u, r, \h{l-1}_v) \in \mathbb{R}^{\widetilde{d}}$ for any edge $e = (u, r, v) \in \gE$;
        \item The \underline{aggregation function} $\widetilde{f}_A$ that computes $\rva^{(l)}_v :=  \widetilde{f}_A(\{\vmu^{(l)}_{(u, r, v)}: (u, r) \in \mathcal{N}_v\}) = \sum_{(u, r) \in \mathcal{N}_v} \vmu^{(l)}_{(u, r, v)}$;
        \item An \underline{update function} $\widetilde{f}_U$ that computes the embeddings of the next layer as $\h{l}_v = \widetilde{f}_U(\rvh^{(l-1)}_v, \rva^{(l)}_v, \rvx_v)$ for all $v \in \gV$.
    \end{enumerate}
\end{definition}

We remark that there still remains a degree of non-uniqueness in Definition \ref{def:mpgnn_layer_canon} in that a linear transformation of $\vmu^{(l)}_{(u, r, v)}$ introduced by $\widetilde{f}_M$ could pass through $\widetilde{f}_A$ and be absorbed into $\widetilde{f}_U$ without changing the composite function $\widetilde{f}_U \circ \widetilde{f}_A \circ \widetilde{f}_M$.  However, this ambiguity is inconsequential for our purposes, and nonlinear transformations cannot analogously pass through $\widetilde{f}_A$ without changing the form of the resulting composite function.

Additionally, although it may superficially appear as though we have lost expressive power in moving to the canonical form, this is actually not the case:

\begin{proposition} \label{prop:canon_form}
    For any $f_U \circ f_A \circ f_M$ adhering to Definition \ref{def:mpgnn_layer}, there exists a canonical form $\widetilde{f}_U \circ \widetilde{f}_A \circ \widetilde{f}_M$ following Definition \ref{def:mpgnn_layer_canon} that provides an arbitrarily close approximation.
\end{proposition}

Appendix \ref{sec:proof_prop1} contains the proof. So indeed, we can work with the canonical form without any appreciable loss of generality.  From here, we will first narrow our scope in Section \ref{sec:message_function_constraint} to reproducing $f_M$ or $\widetilde{f}_M$ using some $\calA[\ell_{low},~\cdot~]$; later in Section \ref{sec:approx_canonical_form} we zoom out and address approximations to the full canonical form.

\subsection{A Priori Constraint Approximating Message Functions $f_M$ or $\widetilde{f}_M$} \label{sec:message_function_constraint}

To begin, we introduce the following definition:

\begin{definition} \label{def:grad_rep_criteria}
We say that a message function $f_M$ satisfies a \underline{gradient representation criteria} if 
\begin{eqnarray}
f_M(\rvh_u, r, \rvh_v) & = & \frac{\partial \zeta(\rvh_u, \rvh_v; r)}{\partial \rvh_v} \nonumber \\
f_M(\rvh_v, r^{-1}, \rvh_u) & = & \frac{\partial \zeta(\rvh_u, \rvh_v; r)}{\partial \rvh_u} 
\end{eqnarray}
for some function $\zeta: \mathbb{R}^{d} \times \mathbb{R}^d \rightarrow \mathbb{R}$ with Lipschitz continuous gradients, where $r^{-1}$ is the inverse relation of $r$.  Moreover, we extend this definition to the canonical form $\widetilde{f}_M$ via
\begin{eqnarray}
\widetilde{f}_M(\rvh_u, r, \rvh_v) &=&  \left.  \frac{\partial \widetilde{\zeta}(\Phi^\top \rvh_u, \rvz; r)}{\partial \rvz} \right|_{\rvz = \Phi^\top \rvh_v} \nonumber \\
\widetilde{f}_M(\rvh_v, r^{-1}, \rvh_u) &=&  \left.  \frac{\partial \widetilde{\zeta}(\rvz, \Phi^\top \rvh_v; r)}{\partial \rvz} \right|_{\rvz = \Phi^\top \rvh_u} 
\end{eqnarray}
for some $\Phi \in \mathbb{R}^{d\times \widetilde{d}}$ and function $\widetilde{\zeta}: \mathbb{R}^{\widetilde{d}} \times \mathbb{R}^{\widetilde{d}} \rightarrow \mathbb{R}$ with Lipschitz continuous gradients.
\end{definition}
This definition describes message functions that can be expressed in terms of the gradient of an energy term, i.e., $\zeta$ or $\widetilde{\zeta}$.  Moreover, if a pair of bidirectional message functions do \textit{not} satisfy this criteria, then they cannot generally be obtained with a gradient-based $\calA$ applied to any possible $\ell_{low}$.  To see this, consider a graph with just two nodes and no node features for simplicity.  Moreover, assume that the aggregation function is identity (there is only one message in either direction) and the update function is given by $f_U(\rvh_v^{(l)},\vmu^{(l)}_{(u, r, v)}) = \h{l-1}_v - \gamma \vmu^{(l)}_{(u, r, v)}$ for some $\gamma > 0$.  We then observe that any possible $\ell_{low}$ can be expressed w.l.o.g.~as $\ell_{low}(\rmH;\rmW,\calG) = \zeta(\rvh_1,\rvh_2;r)$ for some function $\zeta$. 
 Therefore, the only source for producing the two required message functions, meaning for $1 \rightarrow 2$ and $2 \rightarrow 1$ is taking gradients of the first and second arguments of $\zeta$.

\subsection{Approximating the Full Canonical Form via Energy Optimization} \label{sec:approx_canonical_form}

To proceed, we first require an adequately flexible form of energy function with terms that are, in a sense to be formally quantified, sufficiently aligned with the canonical form serving as our target.  Intuitively, there is a trade-off here:  If this energy is too complex, then iterative optimization algorithms applied to it can produce update rules that deviate from Definition \ref{def:mpgnn_layer_canon}, e.g., the update for any given node $v$ could potentially involve all other nodes in the graph, as opposed to merely those in $\calN_v$.  In contrast, if the energy is too narrowly defined, we will be unable to match the expressive power of the composite $\widetilde{f}_U \circ \widetilde{f}_A \circ \widetilde{f}_M$.

Motivated by these considerations, we propose the family of energies given by
\begin{eqnarray}\label{eq:lower_energy_main}
   &&\hspace*{-1.0cm} \ell_{low}(\rmH ; \rmW, \gG ) :=  \\       
   && \sum_{(u, r, v) \in \gE} f\left(\rvh_u, \rvh_v; r \right) + \sum_{v \in \gV} \left[ \kappa(\rvh_v; \rvx_v) + \eta(\rvh_v) \right], \nonumber
\end{eqnarray}
where $f: \sR^{d} \times \sR^{d} \rightarrow \sR$ and $\kappa: \sR^{d} \rightarrow \sR$ are arbitrary differentiable functions, while $\eta: \sR^d \rightarrow \sR$ has no such requirements, e.g., it can be nonsmooth and discontinuous.  Additionally, $\rmW$ refers to all parameters that may be included within $f$, $\kappa$, and $\eta$.

The role of $\eta$ is to allow for the introduction of node-wise constraints.  For instance, if $\eta(\rvh) := \calI_\infty[\rvh < 0]$, meaning a discontinuous indicator function with infinite weight applied element-wise to the entries of $\rvh$ less than zero, then we are effectively enforcing the constraint that node embeddings must be non-negative \citep{yang2021graph}.  To algorithmically handle this possibility, we also require the notion of a proximal operator \citep{parikh2014proximal}, which is defined as
\begin{equation}
        \mathbf{prox}_{\eta, \gamma}(\rvh) := \arg \min_\rvz \left( \eta(\rvz) + \frac{1}{2\gamma} \left\|\rvz - \rvh \right\|^2 \right).
\end{equation}
Proximal operators $\mathbf{prox}_{\eta, \gamma} : \mathbb{R}^{d} \rightarrow \mathbb{R}^d$ of this form are useful for deriving descent algorithms involving constraints or nonsmooth terms such as $\eta$.  We are now prepared to present our main result of this section, with discussion and interpretations to follow:


\begin{proposition} \label{prop:general_mp_representation}
    For any canonical message-passing layer with constituent $\widetilde{f}_U$, $ \widetilde{f}_A$, and $\widetilde{f}_M$ abiding by Definition \ref{def:mpgnn_layer_canon} and $\widetilde{f}_U$ satisfying Definition \ref{def:grad_rep_criteria}, there exists an algorithm $\calA$ and functions $f$, $\kappa$, and $\eta$ defining $\ell_{low}$ from (\ref{eq:lower_energy_main}) such that
    \begin{equation}
     \calA[\ell_{low},~\cdot ~] = \hat{f}_U \circ  \widetilde{f}_A \circ \widetilde{f}_M,
    \end{equation}
    where $\hat{f}_U : \mathbb{R}^d \times \mathbb{R}^{\widetilde{d}} \times \mathbb{R}^d \rightarrow \mathbb{R}^d$ is defined as
    \begin{equation} \label{eq:approx_update_function}
        \hat{f}_U(\rvh,\rva, \rvx) :=  \mathbf{prox}_{\eta,\gamma}\left(\rvh - \gamma \sigma(\rva)\left[\Phi \rva + \kappa'(\rvh;\rvx)\right]  \right)
    \end{equation}
    with $\sigma(\rva) : \mathbb{R}^{\widetilde{d}} \rightarrow \mathbb{R}^+ > 0$ as an arbitrary positive function, $\Phi \in \mathbb{R}^{d \times \widetilde{d}}$ as an arbitrary matrix, and $\kappa'$ as the first-order derivative of $\kappa$.  Moreover, there exists a $\gamma' > 0$ such that for any $\gamma \in (0,\gamma']$, 
    \begin{equation}
        \ell_{low}\left(\gA[\ell_{low}, \rmH] ; \calG\right) ~ \leq ~ \ell_{low}(\rmH; \calG).
    \end{equation}
\end{proposition}

Several salient points are worth highlighting as we unpack this result:
\begin{itemize}
    \item Once we have granted that $\widetilde{f}_U$ adheres to Definition \ref{def:grad_rep_criteria}, all approximation error between the target composite message passing function $\widetilde{f}_U \circ  \widetilde{f}_A \circ \widetilde{f}_M$ and the proposed $\calA[\ell_{low},~\cdot ~]$ is confined to mismatch in the respective update functions $\widetilde{f}_U$ and $\hat{f}_U$, not the message and aggregation functions $\widetilde{f}_M$ and $\widetilde{f}_A$.

    \item  While $\hat{f}_U$ cannot exactly represent all possible $\widetilde{f}_U$, it maintains considerable degrees of freedom for replicating important special cases, including many/most commonly adopted update functions.  This flexibility comes from three primary sources: namely, $\eta$, $\sigma$, and $\kappa$ can all be freely chosen to widen the expressiveness of $\hat{f}_U$; we note that $\eta$ and $\kappa$ are determined by $\ell_{low}$, while $\sigma$ is associated with $\calA$ (see Appendix \ref{sec:proof_prop2} for further details).

    \item The function $\eta$, through its influence on $\mathbf{prox}_{\eta,\gamma}$, can introduce a wide variety of nonlinear activations into the message passing layer.  And in general, since $\eta$ is arbitrary, $\mathbf{prox}_{\eta,\gamma}$ can be any arbitrary proximal operator.  For example, $\mathbf{prox}_{\eta,\gamma}$ can implement any non-decreasing function of the elements of $\rvh$ \citep{gribonval2020characterization}, such as popular ReLU activations among other things.

    \item The function $\sigma$ allows us to introduce softmax self-attention in the message-passing layer.  This capability is exactly analogous to how an optimization perspective can introduce self-attention into typical Transformer layers \citep{yang2022transformers}.  

    \item Meanwhile, the function $\kappa$ provides for arbitrary interactions between node embeddings and input node features.  It also allows us to directly generalize (\ref{eq:simple_mp_case}).  More concretely, when we choose $\eta(\rvh) = 0$ (such that $\mathbf{prox}_{\eta,\gamma}$ degenerates to an identity mapping), $\sigma(\rva) = 1$, and $\kappa'(\rvh; \rvx) = \rvh - \pi(\rvx; \rmW)$, we exactly recover the message passing layer from (\ref{eq:simple_mp_case}).

    \item Increasing the expressiveness of $\hat{f}_U$ to more tightly match an arbitrary $\widetilde{f}_U$ is challenging and possibly infeasible (at least using standard descent methods for $\calA$).  See Appendix \ref{sec: generalizing_expressiveness} for further discussion.
    
\end{itemize}


While the proof of Proposition \ref{prop:general_mp_representation} is deferred to Appendix \ref{sec:proof_prop2}, it is based on assigning $\calA$ to be a form of proximal gradient descent with preconditioning.  Although this class of algorithm has attractive convergence guarantees, there remains potential to improve the convergence rate.  Within the present context, this is tantamount to reducing the actual number of message-passing layers needed to approach a minimum of $\ell_{low}$.  We conclude by noting that it is possible to retain the expressiveness of (\ref{eq:approx_update_function}) and the efficient message passing structure of Definition \ref{def:mpgnn_layer_canon} while accelerating convergence through the use of momentum \citep{polyak1964some} and related \citep{kingma2014adam}; see Appendix \ref{sec:momentum}.

\vspace*{-0.0cm}
\section{On Broader Graph ML Regimes} \label{sec:broader_graph_ML}
\vspace*{-0.2cm}
In the previous section, our focus was on deriving flexible MP-GNN layers obtained by applying an algorithm $\calA$ to the minimization of a lower-level energy  $\ell_{low}$.  Through Proposition \ref{prop:general_mp_representation}, we demonstrated that the operator $\calA[\ell_{low}, ~\cdot~]$ is equivalent to a MP-GNN layer with unrestricted message function $\widetilde{f}_M$, aggregation function $\widetilde{f}_A$, and approximate update function $\hat{f}_U$ as defined by (\ref{eq:approx_update_function}).  Analogous to any other MP-GNN model, these components can be iterated for $L$ steps to produce final embeddings $\rmH^{(L)}$ that can be inserted into the supervised node classification loss as in (\ref{eq:general_bilevel_upper}) to complete an end-to-end bilevel optimization pipeline.

We now expand our scope to encompass other interrelated graph ML domains, building
on the general optimization-based message-passing layers derived in Section \ref{sec:approx_canonical_form}.  Because our motivation stems from the principle that \textit{BiLevel Optimization Offers More Graph Machine Learning}, in the sequel we refer to our framework as \textit{BloomGML}.
\vspace*{-0.2cm}
\subsection{The General BloomGML Framework}
\vspace*{-0.2cm}
We formalize BloomGML via the optimization problem
\vspace*{-0.2cm}
\begin{eqnarray} \label{eq:BloomGML}
    && \hspace*{-0.7cm} \min_{\rmW,\Theta} \ell_{up}\left( \rmH^{(L)}(\rmW),\rmY';\Theta,\gG \right)  \\ 
    && \hspace*{-0.7cm}  \mbox{s.t.~}\rmH^{(l)}   =  \gA\left[\ell_{low}(~\cdot~;\rmW,\gG), \rmH^{(l-1)} \right], \forall l=1,\ldots,L, \nonumber
\end{eqnarray}
where $\calA$ and $\ell_{low}$ follow from Proposition \ref{prop:general_mp_representation}   while $\ell_{up}$ is a generic upper-level loss that can be specified on an application-specific basis, e.g., (\ref{eq:general_bilevel_upper}) is a special case of this form.  Additionally, as before we assume that rows of $\rmY'$ are labels; however, for maximum generality we need not assume that these labels correspond with nodes.  In this way, we can extend BloomGML to generic GNN tasks besides node classification.  More importantly though, we can also form deep connections with non-GNN regimes equally as well.  In this regard, we consider three such examples next: knowledge graph embeddings (KGEs), label propagation (LP), and what are often called graph-MLP models.


\subsection{Knowledge Graph Embedding Models} \label{sec:bloom_for_kge}

\paragraph{Basics.} Knowledge graphs are heterogeneous graphs, typically \textit{without} node features or labels, where each triplet/edge $e = (u,r,v)$ contains relational information between two entities, e.g., $u = $ `Paris', $v=$ `France', and $r=$ `capital of' \citep{ji2021survey}.  The goal of knowledge graph completion (KGC) is to predict unobserved/missing factual triplets given an existing knowledge graph of interest.  A widely adopted approach to this fundamental problem is based on learning knowledge graph embeddings (KGEs) as follows.  First, all existing triplets $\calE$ are treated as positive samples, while a complementary set of negative samples are obtained by randomly choosing triplets $e^- = (u^-,r^-,v^-) \notin \calE$, with $u^-,v^- \in \calV$ and $r^- \in \calR$; we can assemble these in a negative graph $\calG^- = \{\calV,\calR,\calE^- \}$, where $\calE^-$ denotes the set of false triplets. 

We then train a KGE model by minimizing a binary cross-entropy loss of the form
\vspace*{-0.1cm}
\begin{eqnarray} \label{eq:kge_basic_loss}
   \ell_{kge}(\rmH,\rmE) &:=&  \sum_{e \in \calE} \log \rho\left[ s\left(\rvh_u,\rve_r,\rvh_v \right) \right] \\
   &+& \sum_{e' \in \calE^-} \log\left(1- \rho\left[ s\left(\rvh_{u^-},\rve_{r^-},\rvh_{v^-} \right) \right]  \right), \nonumber
\end{eqnarray}
where $\rho$ denotes a standard sigmoid function, $\rmH \in \mathbb{R}^{n\times d}$ are node embeddings as before, and $\rmE \in \mathbb{R}^{m\times d'}$ are embeddings for each relation type $r \in \calR$.  Finally, $s : \mathbb{R}^d \times \mathbb{R}^{d'} \times \mathbb{R}^d \rightarrow \mathbb{R}$ is a so-called score function, which evaluates the compatibility of the triplet embeddings, with a higher score indicating a greater likelihood of being a true triplet \citep{zheng2020dgl}. Additionally, (\ref{eq:kge_basic_loss}) is frequently used as a training loss for traditional heterogeneous GNN models applied to knowledge graphs \citep{schlichtkrull2017modeling}.  Note that at inference time, a candidate triplet can be evaluated by computing the score of the corresponding output-layer embeddings produced by the GNN.


\paragraph{KGEs and \model.}  We now demonstrate how the above process emerges as a special case of \model, and the interpretable implications of this association, in particular relative to GNN modeling.  Let $\bar{\calG} := \calG \cup \calG^- = \{\calV,\bar{\calR},\bar{\calE} \}$, where $\bar{\calE} = \calE \cup \calE^-$.  We also define $\bar{\calR} = \calR \cup \calR^-$ where $\calR^-$ denotes a dummy set of negative relations such that, for every $r \in \calR$ there exists a $r^- \in \calR^-$ representing that an original relation $r$ is now appearing in a negative triple. It follows that $|\bar{\calR}| = 2m$.  Also, we assume that $\rve_r = \rve_{r^-}$ for all $r$, i.e., the relation type embeddings are shared.
For \model, following (\ref{eq:lower_energy_main}) we define $\ell_{low}$ w.r.t.~$\bar{\calG}$ as
\begin{equation}
 \ell_{low}(\rmH ; \rmW,\bar{\calG} ) = \sum_{\bar{e} \in \bar{\calE}} f(\rvh_{\bar{u}}, \rvh_{\bar{v}};\bar{r}),
\end{equation}
where we choose $\rmW = \rmE$ and 
\begin{equation} \label{eq:BloomGML_kge_model}
    f(\rvh_{\bar{u}}, \rvh_{\bar{v}};\bar{r}) = \left\{ \begin{array}{ll}  \log \rho\left[ s\left(\rvh_u,\rve_r,\rvh_v \right) \right], & \hspace*{-0.2cm} \bar{e} \in \calE \\ \log\left(1- \rho\left[ s\left(\rvh_{u^-},\rve_{r^-},\rvh_{v^-} \right) \right]  \right), & \hspace*{-0.2cm} \bar{e} \in \calE^-
    \end{array} \right. 
\end{equation}
while implicitly assuming $\kappa$ and $\eta$ are zero.  By constructing $\ell_{low}$ in this way, the lower-level minimization problem provides optimal node embeddings \textit{conditioned on fixed relation embeddings} $\rmE$.  Hence if we execute a descent algorithm $\calA$ for $L$ steps (e.g., gradient descent), we can obtain some $\rmH^{(L)} \equiv \rmH^{(L)}(\rmW) \equiv \rmH^{(L)}(\rmE)$.

To complete the full \model specification per (\ref{eq:BloomGML}), we must also select $\ell_{up}$. In order to align with KGE learning, we choose $\Theta = \rmW = \rmE$, $\rmY' = \{\emptyset\}$ and define
\begin{eqnarray} \label{eq:L_up_for_KGE}
    \ell_{up}\left( \rmH^{(L)} (\rmW),\rmY';\Theta,\bar{\calG} \right) \hspace*{-0.2cm}& \equiv & \hspace*{-0.2cm}\ell_{up}\left( \rmH^{(L)} (\rmE),\{\emptyset\};\rmE,\bar{\calG} \right) \nonumber \\ & := & \hspace*{-0.2cm}\ell_{kge}\left(\rmH^{(L)}(\rmE),\rmE \right),
\end{eqnarray}
\vspace*{-0.2cm}
where $\ell_{kge}$ follows (\ref{eq:kge_basic_loss}).  

\paragraph{Key Implications of this Reformulation.}  Per Proposition \ref{prop:general_mp_representation} and the downstream \model lens from above, it follows that if we minimize (\ref{eq:L_up_for_KGE}) over just $\rmE$, we are effectively reproducing a minimizer of the original KGE loss from (\ref{eq:kge_basic_loss}), with $\rmH$ computed by $L$ steps of $\calA$.  \textit{This implies that learning KGEs equates to training an} $L$\textit{-layer MP-GNN on graph $\bar{\calG}$ with parameters} $\rmE$.  Of particular relevance here, recent work has been devoted to empirically showing that when training powerful heterogeneous GNN models using (\ref{eq:kge_basic_loss}), contrary to conventional wisdom, completely removing the GNN message-passing layers does \textit{not} actually harm performance.  However, this observation is directly explained by the analysis herein: \textit{Explicit external message-passing is not necessary since we are already training an implicit MP-GNN when solving  (\ref{eq:kge_basic_loss}) over unconstrained embeddings}.  Beyond this finding, \model offers an additional practical benefit.  In brief, because we are free to choose $L$, in inductive settings we can pick a relatively small value such that $\rmE$ is learned to adjust accordingly.  Hence when new nodes are introduced,\textit{ their embeddings can be quickly approximated via only $L$ steps of} $\calA$, rather than full model retraining.  We test this capability in Section \ref{sec:experiments}.

\vspace*{-0.1cm}
\subsection{Other Graph ML Regimes} \label{sec:other_graph_ML}
\vspace*{-0.1cm}
\model elucidates other graph ML scenarios as well, particularly LP and graph-regularized MLP models.  While full details are deferred to Appendix \ref{sec: lp_gmlp}, we note here that LP variants can be recast within \model when we treat observed labels as input features and $\calA$ as proximal gradient descent. We also provide supplementary experiments that verify \model predictions regarding the relationship between LP and special cases of graph-regularized MLPs.





\vspace*{-0.0cm}
\section{Empirical Validation and Analysis} \label{sec:experiments}
\vspace*{-0.1cm}

As \model represents a conceptual framework that has independent value drawn from its explanatory/unifying nature, not just pushing SOTA per se, the motivations for our experiments are two-fold:
\vspace*{-0.3cm}
\begin{enumerate}
\item Showcase the versatility and interpretability of BloomGML across diverse testing regimes;
\item Demonstrate the explanatory power of the bilevel optimization lens applied to graph ML tasks.
\end{enumerate}
\vspace*{-0.2cm}

Despite these quite general aims, with limited space we only include a few representative use cases and coincident analyses.  That being said, our experiments still span quite diverse settings, where SOTA methods are different and often rely on domain-specific architectures and/or heuristics. Hence our strategy is simply to pick strong, domain-specific baselines for comparisons across each task and narrative purpose, noting that these baselines generally outperform standard/generic alternatives (on a task-by-task basis) per results in prior work.  Additionally, throughout this section, we use \textit{BloomGML} to describe models derived from (\ref{eq:BloomGML}), with differentiating descriptions accompanying each experiment. Also, please see Appendix \ref{sec: exper_details} for comprehensive details of experimental setups as well as reproduction of result tables with error bars.

\vspace*{-0.1cm}
\paragraph{Mitigating Spurious Input Features.} As motivated in Section \ref{sec:bilevel_example}, bilevel optimization with
an appropriate $\ell_{low}$ should in principle be able to localize spurious input features, allowing the model to focus on more informative network effects when needed.  However, to our knowledge, this capability has not as of yet been exploited.  To this end, we compare the original base model formed via (\ref{eq:quad_example_energy}) and (\ref{eq:simple_mp_case}), with a more robust version drawn from the general \model framework of (\ref{eq:BloomGML}).  In particular, since quadratic regularizers can be sensitive to outliers, we modify (\ref{eq:quad_example_energy}) by choosing $\kappa(\rvh; \rvx) =  \delta\left(\rvh - \pi(\rvx;\rmW)\right) $ within \model, where $\delta : \mathbb{R}^d \rightarrow \mathbb{R}^+$ represents a robust Huber loss penalty \citep{huber1992robust} applied to each input dimension and summed.  We then conduct node classification experiments over datasets Cora, Citeseer, Pubmed, and ogbn-arxiv by randomly corrupting 20\% of node features prior to training.  Results are displayed in Table \ref{tab: noise} where multiple conclusions are evident: The Huber loss improves the accuracy across all datasets, while correctly identifying the majority of outliers.  


\begin{table}[t]
\captionsetup{font=footnotesize}
\footnotesize
\caption{\textit{Performance mitigating spurious input features.} Here the detect ratio is obtained by computing $\|\rvh_v - \pi(\rvx_v;\rmW)  \|_2$ for all $v\in \calV$ and then segmenting out the percentage of corrupted nodes within the largest 20\%.}
\vspace{-0.3cm}
\label{tab: noise}
\centering
\resizebox{\columnwidth}{!}{
\begin{tabular}{cc|cccc}
\specialrule{.15em}{.05em}{.05em}
&   & Cora  & Citeseer & Pubmed & Arxiv \\ \hline
\multirow{2}{*}{Accuracy}& GCN & 51.40 & 46.40 & 59.00 & 64.03 \\ 
& Base from (\ref{eq:quad_example_energy})                & 54.97  &   38.98     & 45.67   &     47.63   \\
& \model    & \textbf{65.83 } &  \textbf{49.33 }    & \textbf{72.70}  & \textbf{69.11 }   \\ \hline
Detect Ratio & \model & 94.27 & 87.82 & 96.98 & 100.00\\
\specialrule{.15em}{.05em}{.05em}
\end{tabular}
}
\vspace*{-0.1cm}
\end{table}

\begin{table}[h]
\captionsetup{font=footnotesize}
\footnotesize
\caption{\textit{Node classification on  heterophily graphs}.
Following convention, accuracy (\%) is reported for Roman and Amazon, while ROC-AUC (\%) is used for Minesweeper, Tolokers, and Questions.}
\vspace{-0.3cm}
\label{tab: heter}
\resizebox{\columnwidth}{!}{
\begin{tabular}{l|ccccc|c}
\specialrule{.15em}{.05em}{.05em}
& Roman     & Amazon   & Minesweep      & Tolokers         & Questions & Avg.        \\ \hline
FAGCN   &65.22 & 44.12 & 88.17 & 77.75 & 77.24 & 70.50 \\
FSGNN      & 79.92  & 52.74  & 90.08  &  82.76  & \textbf{78.86 } & 76.87\\
GBK-GNN       & 74.57   & 45.98 & 90.85  & 81.01  & 74.47  & 73.38\\ 
JacobiConv      & 71.14 &  43.55  & 89.66  &  68.66 & 73.88  & 69.38\\ \hline
Base from (\ref{eq:quad_example_energy})      & 76.63  & 52.37  & 88.97 & 80.91 & 76.72 & 75.12  \\
\model & 84.45 & \textbf{52.92 } & 91.83 & 84.84 & 77.62 & 78.33 \\
\model (w/Hub.) & \textbf{85.26} & 51.00 & \textbf{93.30} & \textbf{85.92} & 77.93 & \textbf{78.68} \\
\specialrule{.15em}{.05em}{.05em}
\end{tabular}
}
\vspace*{-0.3cm}
\end{table}

\vspace*{-0.1cm}
\paragraph{Comparison on Heterophily Graphs.} For graphs exhibiting heterophily \citep{zhu2021graph}, nodes sharing an edge are less likely to have the same label.  Hence an energy function that pushes the respective node embeddings to a common value may be counterproductive, i.e., as in the edge-dependent term from (\ref{eq:quad_example_energy}).  Instead, to address heterophily graphs using BloomGML, we form $\ell_{low}$ via $f(\rvh_v,\rvh_u;r) = \left\|\rvh_u \rmC - \rvh_v \right\|_{2}^2$, where $\rmC \in \mathbb{R}^{d\times d}$ is a trainable weight and $r$ is ignored for the homogeneous case.  Additionally, to reduce sensitivity to less reliable node features that could potentially accompany heterophily graphs, we consider $\kappa(\rvh; \rvx) =  \delta\left(\rvh - \pi(\rvx;\rmW) \right)$ to contrast with $\kappa(\rvh; \rvx) =  \| \rvh - \pi(\rvx;\rmW) \|^2_2$.  Finally, we select $\eta(\rvh) = \calI_\infty[\rvh < 0]$ and $\sigma(\rva) = 1$ and apply proximal gradient descent for $\calA$.   Table \ref{tab: heter} displays results using the recent heterophily benchmarks Roman, Amazon, Minesweeper, Tolokers, and Questions from \cite{platonov2023a}.  We compare BloomGML against recent GNN models explicitly designed for handling heterophily, namely, FAGCN~\citep{bo2021beyond}, FSGNN \citep{maurya2022}, GBK-GNN \citep{du2022gbk}, and JacobiConv \citep{wang2022powerful}.  For reference, we also compare with the bilevel baseline formed from (\ref{eq:quad_example_energy}); additional GNN baselines from \cite{platonov2023a} are discussed in Appendix \ref{sec: exper_details}.  On average, we observe that BloomGML achieves the best accuracy here.  Even so, our focus is not on solving heterophily problems per se, but rather, demonstrating that interpretable modifications to $\ell_{low}$ can induce predictable effects.

\begin{figure}[t]
\captionsetup{font=footnotesize}
    \centering
    \includegraphics[width=0.38\textwidth]{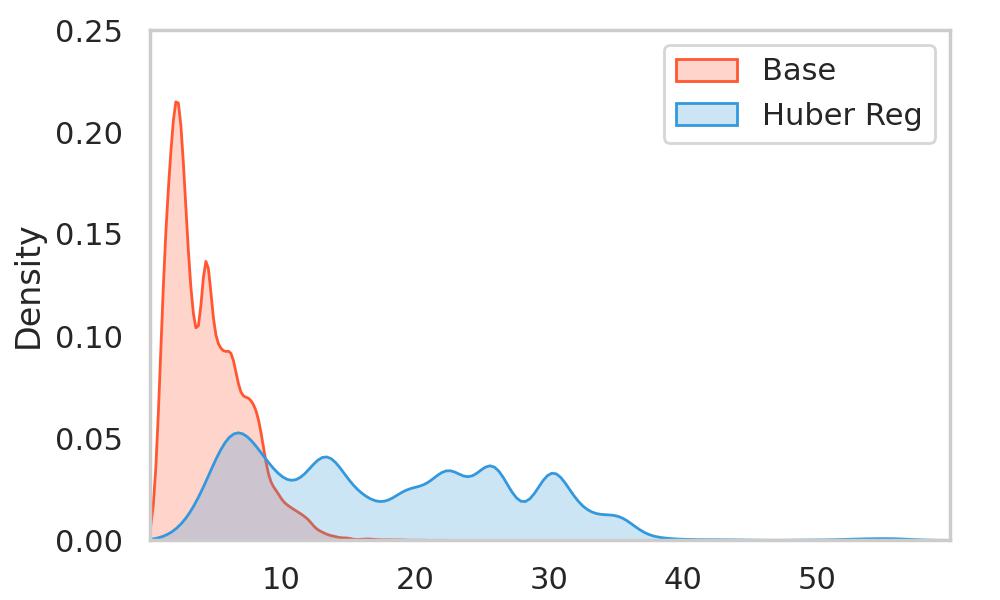}
    \vspace{-0.3cm}
    \caption{\textit{$\|\h{L}_u - \h{L}_v\|_2$ density for ${(u, v)\in\gE}$ but with different labels in the Roman dataset. 
  } }
  \label{fig:norm_dist}
  \vspace{-0.3cm}
\end{figure}

Regarding the latter, we present one additional visualization afforded by the \model paradigm.  On the Roman dataset, we incorporate $\delta$ within the function $f$, which allows greater freedom for embeddings sharing an edge but with different labels to deviate from one another.  The distribution of $\|\h{L}_u - \h{L}_v\|_2$ for ${(u, v)\in\gE}$ but $\rvy_v \neq \rvy_u$ (i.e., different labels) is shown in Figure \ref{fig:norm_dist}.  Clearly, the Huber loss allows greater deviation in embeddings sharing an edge as expected.  Interestingly, the accuracy using Huber within $f$ increases to 88.12\%.


\begin{figure}[t]
\captionsetup{font=footnotesize}
\setlength{\tabcolsep}{3pt}
\footnotesize
\centering
\captionof{table}{\textit{Node classification accuracy involving heterogeneous graphs.} \model is modified from HALO to include robust regularization within $\kappa(\rvh; \rvx)$.}
\vspace{-0.2cm}
\label{tab: halo}
\begin{tabular}{c|cccc}
\specialrule{.15em}{.05em}{.05em}
  & AIFB  & MUTAG & BGS   & AM    \\ \hline
HALO     & 97.2 & 80.9 & 89.7 & 85.9 \\
\model    & 97.2 & \textbf{82.4} & \textbf{93.1} & \textbf{86.9} \\ 
\specialrule{.15em}{.05em}{.05em}
\end{tabular}
\end{figure}

\paragraph{Comparison on Heterogeneous Graphs.}  The base model formed via (\ref{eq:quad_example_energy}) has also recently been extended to heterogeneous graphs via the HALO model \citep{ahn2022descent}.  However, this extension remains limited in its strict adherence to the quadratic edge- and input feature-dependent penalty terms as in the original (\ref{eq:quad_example_energy}).  To this end, we modify HALO within the confines of the \model framework to explore non-quadratic penalties (Log-Cosh 
and Huber) while maintaining equivalence across all other aspects of model design.  Node classification results are in Table \ref{tab: halo}, where we observe that all else being equal, these changes lead to an improvement in accuracy over HALO.

\paragraph{Knowledge Graph Completion.}  As mentioned at the end of Section \ref{sec:bloom_for_kge}, \model is particularly well-suited for inductive KGE tasks such as KGC.  To this end, we analytically and empirically compare against the recent RefactorGNN model \citep{chen2022refactor}, which includes related analysis showing how a particular score function, DistMult \citep{yang2014embedding}, when combined with a softmax training loss over positive samples, has some commonalities with GNN layer structure.  From a practical standpoint, RefactorGNN involves training a KGE model while resetting node embeddings to initial values in regular intervals to facilitate the inductive setting (the edge embeddings are trained as usual).  While certainly a noteworthy contribution, there nonetheless remain three limitations of RefactorGNN relative to \model.

First, the RefactorGNN model does not actually induce an efficient/sparse MP-GNN, as there remains a global hub node to which all other nodes are connected and receive messages.  Secondly, the supporting analysis provided in \cite{chen2022refactor} only directly addresses the softmax/DistMult pairing while excluding consideration of negative samples; in contrast, \model serves as a general framework covering a wide variety of both KGE and non-KGE models alike, and within which we have explicitly accounted for negative samples through $\bar{\calG}$.  And lastly, the repeated reinitialization of node embeddings during RefactorGNN training can be viewed as a heuristic, with no guarantee of convergence; \model sidesteps this issue altogether within an integrated bilevel optimization framework.

To conduct empirical comparisons with RefactorGNN, we must first specify (\ref{eq:lower_energy_main}) for \model.  We take inspiration from NBFNet \citep{zhu2021neural} and design $\elow$ such that one step of proximal gradient descent mimics an NBFNet model layer; see Appendix \ref{sec: nbfnet} for derivations showing how this process aligns with an MP-GNN layer when incorporated within \model under the right circumstances.  We also set $\kappa(\rvh;\rvx) = \eta(\rvh) = 0$, $\sigma(\rva) = 1$ for simplicity.  Inductive KGC results are shown in Table \ref{tab: kgc} on WordNet18RR\_v1 and FB15K237\_v1 datasets \citep{teru2020inductive}.  Across different score functions TransE and DistMult, and aggregators Sum and PNA \citep{corso2020principal}, \model achieves strong performance relative to RefactorGNN.  Note that in Table \ref{tab: kgc} we have reported the best RefactorGNN model from \cite{chen2022refactor} as there is presently no publicly-available code for reproducibiliy.

\begin{table}[t]
\captionsetup{font=footnotesize}
\footnotesize
\centering
\caption{\textit{Inductive KGC tasks.} Reported results based on the Test Hits@10 metric. Sum and PNA are aggregators while T and D are TransE and DistMult score functions, respectively.}
\vspace{-0.2cm}
\label{tab: kgc}
\resizebox{\columnwidth}{!}{
\begin{tabular}{l|cccc|cccc}
\specialrule{.15em}{.05em}{.05em}
       & \multicolumn{4}{c}{WN18RR\_v1}                                & \multicolumn{4}{c}{FB15k-237\_v1}                             \\
         & PNA(T)          & Sum(T)         & PNA(D)           & Sum(D)          & PNA(T)          & Sum(T)         & PNA(D)           & Sum(D)          \\
\hline
RefactorGNN     &    $\slash$      &     $\slash$    &    $\slash$      & 0.885       &    $\slash$     &   $\slash$     &     $\slash$     & 0.787        \\
\model  & 0.952        & 0.937       & \textbf{0.960}         & 0.946        & 0.744        & 0.747       & \textbf{0.836}         & 0.792   \\
\specialrule{.15em}{.05em}{.05em}
\end{tabular}
}
\end{table}

\begin{figure}[t]
\captionsetup{font=footnotesize}
    \centering
    \includegraphics[width=0.4\textwidth]{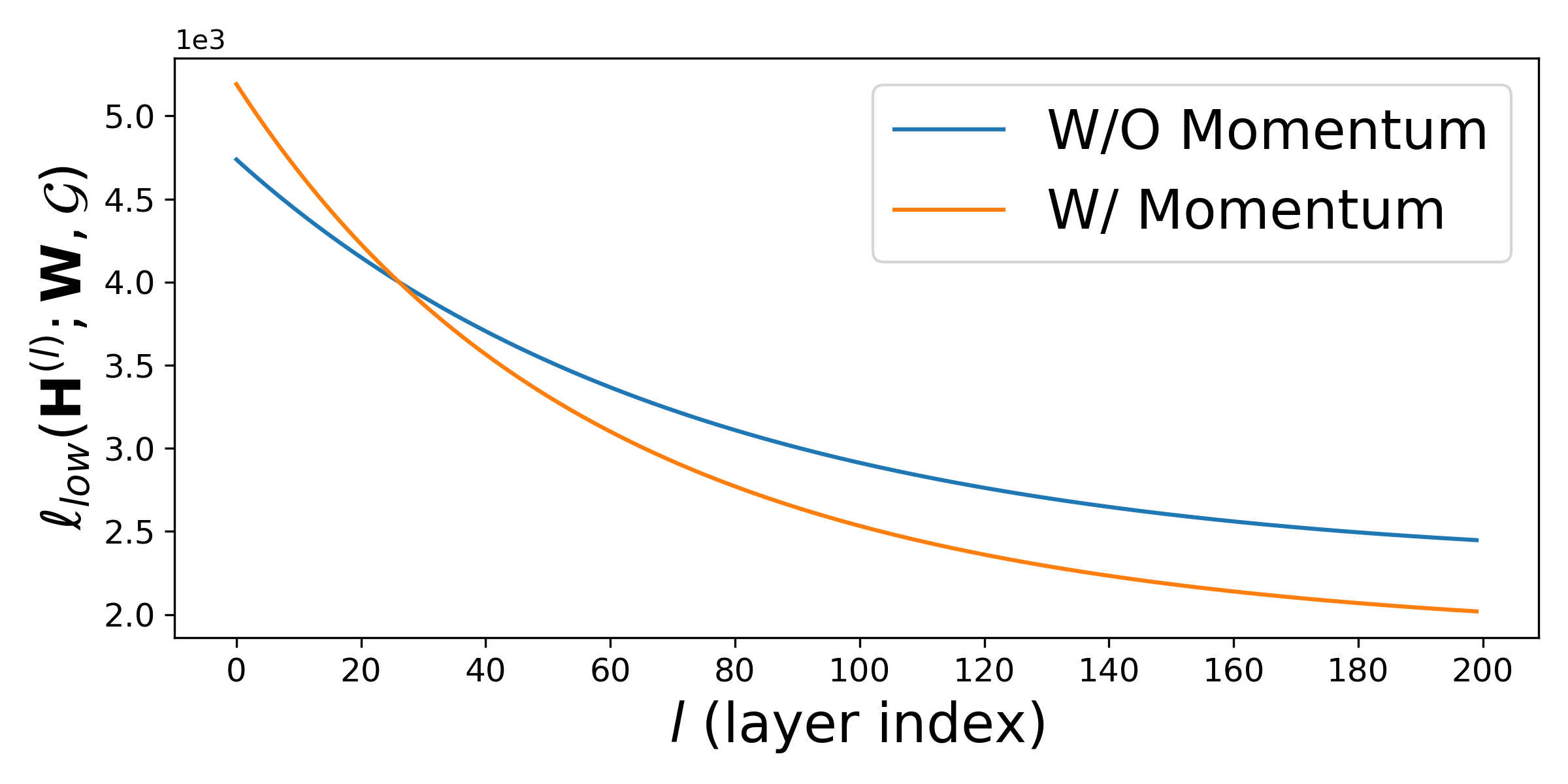} 
     \vspace{-0.4cm}
    \caption{$\elow$ values versus propagation steps on Cora.}
    \label{fig: energy}
    \vspace*{-0.3cm}
\end{figure}

\vspace*{-0.1cm}
\paragraph{Incorporation of Momentum.}  We demonstrate how the incorporation of momentum within $\calA$, as proposed at the end of Section \ref{sec:approx_canonical_form} and detailed in Appendix \ref{sec:momentum}, can maintain MP-GNN structure while expediting convergence.  Results are shown in Figure \ref{fig: energy} and Table \ref{tab: momentum} using a simple base model akin to (\ref{eq:quad_example_energy}), where we observe that momentum can potentially improve \model performance by reducing $\ell_{low}$ more quickly.

\begin{table}[h]
\captionsetup{font=footnotesize}
    \setlength{\tabcolsep}{3pt}
    \footnotesize
    \centering
    \captionof{table}{\textit{Node classification with momentum.} The optimization algorithm for \model is SGD with momentum.}
\vspace{-0.3cm}
\label{tab: momentum}
\centering
\begin{tabular}{c|cccc}
\specialrule{.15em}{.05em}{.05em}
              & Cora  & Citeseer & Pubmed & Arxiv \\ \hline
SGD                 & 80.1 &  73.2    & 78.0  &     72.0    \\
\model               & \textbf{83.4} &  \textbf{74.0}    & \textbf{80.7}  &     \textbf{72.6}    \\
\specialrule{.15em}{.05em}{.05em}
\end{tabular}
\vspace*{-0.2cm}
\end{table}

\paragraph{Additional Results.}  Finally, please see Appendix \ref{sec:additional_experiments} for  additional experiments including an ablation over learning $\lambda$, the trade-off parameter from (\ref{eq:quad_example_energy}), as well as further demonstration of the versatility of \model.

\vspace*{-0.1cm}
\section{Conclusion}
\vspace*{-0.2cm}
In this work we have introduced the \model framework, which provides a novel lens for understanding various graph ML paradigms and introducing interpretable, targeted enhancements.  \textit{Let graph ML bloom}.

\newpage
\bibliography{ref,reference2,wipf_refs}

\newpage

\onecolumn
\appendix




\section{Appendix Overview}

Appendix content more-or-less follows the order in which it was originally referenced in the main text; we summarize as follows:

\begin{itemize}
    \item In Section~\ref{sec:derive_mp_special_case}, we derive the the message passing special case from~(\ref{eq:simple_mp_case}).
    \item In Section~\ref{sec: generalizing_expressiveness}, we discuss why increasing the expressiveness of \model in meaningful ways is challenging.
    
\item In Sections~\ref{sec:proof_prop1} and \ref{sec:proof_prop2}, we provide the proofs of Propositions~\ref{prop:canon_form} and~\ref{prop:general_mp_representation}, respectively.
    
    \item In Section~\ref{sec:momentum}, we present a general framework for accelerating convergence and related special cases.
    \item In Section~\ref{sec: lp_gmlp}, we fill in details relating \model to label propagation and graph-regularized MLPs. We also provide supporting empirical results. 
    \item In Section~\ref{sec: exper_details}, we provide model implementation and experiment details related to Section~\ref{sec:experiments}.
    \item In Section~\ref{sec: nbfnet}, we explore the connection between NBFNet and \model.
    \item In Section~\ref{sec:additional_experiments}, we present additional empirical results/ablations.
    
\end{itemize}


\section{Derivation of Message Passing Special Case from (\ref{eq:simple_mp_case})}  \label{sec:derive_mp_special_case}

The message passing updates from (\ref{eq:simple_mp_case}) in Section \ref{sec:bilevel_example} are based on the energy function from (\ref{eq:quad_example_energy}), which we repeat here for convenience as 
\begin{equation}
    \elow(\rmH; \rmW, \gG) = \sum_{v \in \gV} \tfrac{1}{2}\left\|\rvh_v - \pi(\rvx_v;\rmW)  \right\|_{2}^2 + \tfrac{\lambda}{2} \sum_{(u,v) \in \gE} \left\|\rvh_u - \rvh_v \right\|_{2}^2.
\end{equation}
Taking the derivative of this $\elow$ w.r.t.~$\rvh_v$ we obtain
\begin{equation}
    \nabla_{\rvh_v} \elow(\rmH; \rmW, \gG) = \rvh_v - \pi(\rvx_v; \rmW) - \lambda \sum_{u \in \gN_v} (\rvh_u - \rvh_v).
\end{equation}
Let $\vmu^{(l)}_{(u, r, v)} := \rvh^{(l-1)}_u - \rvh^{(l-1)}_v$ and $\rva_v^{(l)} := \sum_{u \in \gN_v} \vmu^{(l)}_{(u, r, v)}$. There will then exist an update function $f_U(\rvh^{(l-1)}_v,\rva^{(l)}_v,\rvx_v)$ such that
we have
\begin{equation}
    \rvh^{(l)}_v = \h{l-1}_v - \gamma \nabla_{\rvh_v} \elow(\rmH; \rmW, \gG) = f_U(\rvh^{(l-1)}_v,\rva^{(l)}_v,\rvx_v) := (1-\gamma) \rvh_v^{(l-1)}  + \gamma \lambda  \va^{(l)}_v + \gamma \pi(\rvx_v;\rmW),
\end{equation}
which reproduces (\ref{eq:simple_mp_case}).

\section{Possibilities for Generalizing \model Expressiveness} \label{sec: generalizing_expressiveness}

As alluded to in Section \ref{sec:approx_canonical_form}, increasing the expressiveness of \model in meaningful ways is challenging.  At a trivial level, we could always generalize (\ref{eq:lower_energy_main}) by allowing $f \rightarrow f_{(u,r,v)}$ to vary across each triplet; analogously we could allow $\kappa \rightarrow \kappa_{(v)}$ and $\eta  \rightarrow \eta_{(v)}$ or related.  But for more interesting enhancements, we might consider flexible expressions of permutation-invariant functions given that the input graph is invariant to the order of nodes and triplets.  For example, results from \cite{zaheer2017deep} show that, granted certain stipulations on the input domain, permutation invariant functions operating on a set $\calX$ can be expressed as $\rho[\sum_{x \in \calX} \phi(x)]$ when granted suitable transforms $\rho$ and $\phi$.  From this type of result, it is natural to consider, for example, generalizing the first term of (\ref{eq:lower_energy_main}) via
\begin{equation}
 \sum_{(u, r, v) \in \gE} f\left(\rvh_u, \rvh_v; r \right) \rightarrow \rho\left[ \sum_{(u, r, v) \in \gE} f\left(\rvh_u, \rvh_v; r \right)  \right].
\end{equation}
The problem here though is that, if we take the gradient of this revised term w.r.t.~the node embeddings, by the chain rule the result can depend on all triplets in the graph, which violates the message-passing schema from either Definition \ref{def:mpgnn_layer} or \ref{def:mpgnn_layer_canon}.

\section{Proof of Proposition \ref{prop:canon_form}} \label{sec:proof_prop1}


The basic strategy here will be to push components of the original aggregation function $f_A$ into the revised message and update functions $\widetilde{f}_M$ and $\widetilde{f}_U$, respectively.  And the underlying goal is to do so without losing any expressiveness in the resulting composite function, even while reducing the revised aggregation function $\widetilde{f}_A$ to simple, parameter-free addition.  

To proceed, we rely on a suitable decomposition of permutation-invariant set functions as required by $f_A$ over input messages from $\gN_v$, where the domain of each message is $\mathbb{R}^d$.  In this regard, a number of permutation-invariant function decompositions have been proposed in \cite{zaheer2017deep} with conditions dependent on the domain of the input sets.  The strongest result they present provides a useful general form of exact decomposition; however, unfortunately the underpinning guarantee only holds for input sets drawn from a \textit{countable} universe, which $\mathbb{R}^d$ is of course not.

To work around this limitation, we first assume that $|\calN_v| = p < n$ for all $v \in \calV$, where $p > 0$ is a fixed constant.  We may then apply Theorem 9 from \cite{zaheer2017deep} which stipulates that any continuous $f_A$ defined on a compact set within $\mathbb{R}^d$ can be approximated arbitrarily closely via a decomposition of the form
\begin{equation}\label{eq:aggr_approx}
    f_A\left(\{\vmu^{(l)}_{(u, r, v)}: (u, r) \in \mathcal{N}_v\} \right) \approx \rho\left( \sum_{(u, r) \in \mathcal{N}_v} \phi(\vmu^{(l)}_{(u, r, v)}) \right),
\end{equation}
where $\rho: \sR^{\Tilde{d}} \rightarrow \sR^{d}$ and $\phi: \sR^{d} \rightarrow \sR^{\Tilde{d}}$ are suitable transforms.  

To extend to the general case where $|\calN_v|$ may vary across nodes (since graphs are generally not limited to a fixed node degree), we add $\max_{v \in \calV}|\gN_v|  - |\gN_v|$ dummy neighbors $(u', r')$ to $\gN_v$ for every $v \in \gV$, and form the augmented neighbor sets $\gN'_v$, with the restriction that $\phi(\vmu_{u', r', v}) = 0$. Note that in the neighborhood of the associated dummy messages $\vmu_{(u', r', v)}$, we can no longer guarantee a close approximation to the true function; however, we can cluster such dummy messages in an area of arbitrarily small measure, so their impact is negligible.  Given $\gN'_v$ so defined, $f_A$ can be approximated arbitrarily closely  (except for within a negligible, vanishingly small area of the input domain)  via a decomposition of the form 
\begin{equation}\label{eq:aggr_approx}
    f_A(\{\vmu^{(l)}_{(u, r, v)}: (u, r) \in \mathcal{N}_v\}) \approx \rho\left( \sum_{(u, r) \in \mathcal{N}'_v} \phi(\vmu^{(l)}_{(u, r, v)}) \right) \equiv \rho\left( \sum_{(u, r) \in \mathcal{N}_v} \phi(\vmu^{(l)}_{(u, r, v)}) \right).
\end{equation}
Now define $\widetilde f_M := \phi \circ f_M$, ~~$\widetilde \vmu^{(l)}_{(u, r, v)}:= \widetilde{f}_M(\h{l-1}_u, r, \h{l-1}_v) \in \mathbb{R}^{\widetilde{d}}$,~~ and $\widetilde f_A(\{\widetilde\vmu^{(l)}_{(u, r, v)}: (u, r) \in \gN_v\}) = \sum_{(u, r) \in \mathcal{N}_v} \widetilde \vmu^{(l)}_{(u, r, v)}$.  Then we have 
\begin{equation}
    \rho \circ \widetilde f_A \circ \widetilde f_M \approx f_A \circ f_M
\end{equation}
Furthermore, if we define $\widetilde f_U  := f_U \circ \rho$, we may conclude that
\begin{equation}
    \widetilde{f}_U \circ \widetilde{f}_A \circ \widetilde{f}_M~~\approx~~ f_U \circ f_A \circ f_M
\end{equation}
in the sense described above.

\section{Proof of Proposition~\ref{prop:general_mp_representation}} \label{sec:proof_prop2}

We assume that $\widetilde{f}_M$ satisfies the gradient representation criteria from Definition \ref{def:grad_rep_criteria}. Now recall that the energy function $\elow$ is given by
\begin{equation}
    \ell_{low}(\rmH ; \rmW, \gG ) :=\sum_{(u, r, v) \in \gE} f\left(\rvh_u, \rvh_v; r \right) + \sum_{v \in \gV} \left[ \kappa(\rvh_v; \rvx_v) + \eta(\rvh_v) \right].
\end{equation}
First, we choose 
\begin{equation}
f\left(\rvh_u, \rvh_v; r \right) ~~:= ~~ \widetilde{\zeta}(\Phi^\top \rvh_u, \Phi^\top \rvh_v; r) ,
\end{equation}
where $\Phi$ and $\widetilde{\zeta}$ instantiate the gradient representation criteria for $\widetilde{f}_M$.  From this it follows that
\begin{eqnarray}
 \frac{\partial f\left(\rvh_u, \rvh_v; r \right)}{\partial \rvh_v}   & = & \Phi \widetilde{f}_M(\rvh_u, r, \rvh_v) \nonumber \\
  \frac{\partial f\left(\rvh_u, \rvh_v; r \right)}{\partial \rvh_u}   & = & \Phi \widetilde{f}_M(\rvh_v, r^{-1}, \rvh_u).
\end{eqnarray}
We may then choose $\vmu^{(l)}_{(u, r, v)} := \widetilde{f}_M(\h{l-1}_u, r, \h{l-1}_v) \in \mathbb{R}^{\widetilde{d}}$, and by adding up all gradient terms depending on a given node $v$, we can produce the aggregation function
$\va^{(l)}_v :=  \widetilde{f}_A(\{\vmu^{(l)}_{(u, r, v)}: (u, r) \in \mathcal{N}_v\}) = \sum_{(u, r) \in \mathcal{N}_v} \vmu^{(l)}_{(u, r, v)}$, noting that in certain cases $r = \bar{r}^{-1}$ for some other relation $\bar{r} \in \calR$.  And by linearity of the aggregation, we can later push $\Phi$ through to the update function (see below).

In this way, the gradient of $\ell_{low}$ w.r.t.~$\h{l-1}_v$, excluding the (possibly) non-differentiable $\eta$ term is given by $\Phi \va^{(l)}_v + \kappa'(\h{l-1}_v; \rvx_v)$.  Given the Lipschitz continuous gradients of $\widetilde{\zeta}$, it follows that for a suitably small $\gamma' > 0$, there will always exist an upper bound of the form
\begin{equation}
\ell_{low}(\rmH;\rmW,\calG) \leq \sum_{v \in \calV} \tfrac{1}{2\gamma \sigma(\va_v^{(l)})}\left[\left\|\rvh_v - \left(\h{l-1}_v - \gamma \sigma(\va^{(l)}_v) \left[ \Phi \va^{(l)}_v + \kappa'(\h{l-1}_v; \rvx_v)\right] \right)  \right\|_2^2 +  \eta(\rvh_v)  \right]  + C
\end{equation}
for all $\gamma \in (0,\gamma']$, where $C$ is an irrelevant constant, $\sigma$ is a preconditioner function, and we achieve equality iff $\rmH = \rmH^{(l-1)}$.  From this bound, we can construct an update function by taking a proximal step that is guaranteed to reduce or leave unchanged $\ell_{low}$.  The form is given by
\begin{equation}
    \hat{f}(\rvh^{(l-1)}_v, \va^{(l)}_v, \rvx_v) := \mathbf{prox}_{\eta, \gamma\sigma(\va_v^{(l)})} \left(\h{l-1}_v - \gamma \sigma(\va^{(l)}_v) \left[ \Phi \va^{(l)}_v + \kappa'(\h{l-1}_v; \rvx_v)\right] \right),
\end{equation}
completing the proof.

\section{Accelerating \model Convergence with Momentum/Hidden States} \label{sec:momentum}

At the end of Section \ref{sec:approx_canonical_form}, we briefly alluded to retaining the expressiveness of (\ref{eq:approx_update_function}) and the efficient message passing structure of Definition \ref{def:mpgnn_layer_canon} while accelerating convergence, meaning a smaller value of $L$ will be adequate for approximating a minimum of $\ell_{low}$.  We fill in further details here.  At a high level, this goal can be accomplished through the use of additional hidden states  $\rmS = \{\rvs_v\}_{v \in \gV} \in \sR^{n \times d}$ that instantiate momentum \citep{polyak1964some}.  The latter can mute the amplitude of oscillations in noisy gradients while traversing flatter regions of the loss surface more quickly. A revised version of (\ref{eq:approx_update_function}) festooned with momentum is given by 
\begin{eqnarray}
    \label{eq:momentum_update_fuction}
           && \hspace*{-0.5cm}  \rvs^{(l)} = \beta \rvs^{(l-1)} + (1-\beta)\sigma(\rva^{(l)})\left[\Phi \rva^{(l)} + \kappa'(\rvh^{(l-1)};\rvx)\right] \nonumber \\
           && \hspace*{-0.5cm}  \h{l} = \mathbf{prox}_{\eta, \gamma}\left(\h{l-1} - \gamma \rvs^{(l)} \right),
\end{eqnarray}
where $\beta$ is an additional trade-off parameter. 

Beyond the vanilla version of momentum given by (\ref{eq:momentum_update_fuction}), for reference we now define a more general formulation that encompasses many well-known acceleration methods including AdaGrad~\citep{duchi2011adaptive}, RMSProp~\citep{hinton2012neural}, and Adam~\citep{kingma2014adam} as special cases. 

To begin, let $\elow$ be any differentiable energy function expressible as (\ref{eq:lower_energy_main}) with $\eta = 0$ (we omit handling the more general case for brevity, although it can be derived in a similar way) and $\rvg^{(l)}_v := \Phi \va^{(l)}_v + \kappa'(\h{l-1}_v; \rvx_v) \in  \sR^{d}$. Given auxiliary variables $\rmS = \{\rvs_v\}_{v \in \gV} \in \sR^{n \times d}$ and a coefficient $\beta > 0$, the iterative mapping\footnote{We omit a subscript $v$ in the following context for simplicity.}
\begin{equation}
\label{eq:hidden_state}
    \begin{aligned}
        \rvs^{(l)} &= \beta \rvs^{(l-1)} +  (1 - \beta) \xi(\rvg^{(l-1)}) \\
        \h{l} &= \h{l-1} - \gamma \sigma(\rvs^{(l)}) \varphi(\rvg^{(l-1)})
    \end{aligned}
\end{equation}
provides a general framework for accelerating convergence.  Here $\xi: \sR^{d} \rightarrow \sR^{d}$, and $\varphi: \sR^{d} \rightarrow \sR^{d}$ are linear node-wise functions defined on $\rvg$. $\sigma(\cdot) : \mathbb{R}^{d} \rightarrow \mathbb{R}^+ > 0$ is an arbitrary positive function. Note that when $k > 1$ auxiliary variables are involved, $\varphi$ is extended to $\sR^{kd} \rightarrow \sR^{d}$.  Such an iterative mapping preserves the properties of message-passing functions. To clarify this, let $f_{U'}(\rvh, \rva, \rvs, \rvx)$ denote the mapping defined in (\ref{eq:hidden_state}). For any $v \in \gV$,  we have $\rvh^{(l)}_v = f_{U'}(\rvh^{(l-1)}_v, \rva^{(l)}_v, \rvs^{(l-1)}_v, \rvx_v)$, which means one update step only takes information from $\{v \cup \gN_v\}$. Also, $\rva_v = \sum_{(u, r) \in \gN_v} \vmu_{(u, r, v)}$ is a permutation-invariant function by construction, which means $f_{U'}$ is also permutation invariant over the sets of 1-hop neighbors.

Turning now to the remainder of this section, we verify that the vanilla momentum from above, as well as AdaGrad, RMSProp, and Adam are all special cases of (\ref{eq:hidden_state}).

\paragraph{Momentum as Special Case.}
\label{sec: momentum}

To show that vanilla momentum is a message-passing layer formed as a special case of (\ref{eq:hidden_state}), we let $\xi(\rvg) = \rvg$, $\sigma(\rvs) = \rvs$, and $\varphi = 1$, then the mapping is given by
\begin{equation}
    \begin{split}
        \rvs^{(l)} &= \beta \rvs^{(l-1)} + (1 - \beta) \rvg^{(l-1)}, \\
        \h{l} &= \h{l-1} - \gamma \rvs^{(l)}.
    \end{split}
\end{equation}
which reproduces (\ref{eq:momentum_update_fuction}). 

\paragraph{AdaGrad.}
\label{sec: adagrad}

Here we show AdaGrad~\citep{duchi2011adaptive} is also a special case of~(\ref{eq:hidden_state}). Let $\beta = 0$, $\xi(\rvg) = \mathrm{diag}(\rvg \rvg^T)$, $\varphi(\rvg) = \rvg$, 
and $\sigma(\rvs) = \left(\rvs^{(l)}+\epsilon \right)^{-\frac{1}{2}}$ where $\epsilon$ is a small constant to avoid division by zero.  Then the updating rule of AdaGrad is given by
\begin{equation}
    \begin{split}
        \rvs^{(l)} &= \mathrm{diag}(\rvg^{(l-1)} (\rvg^{(l-1)})^T), \\
        \h{l} &= \h{l-1} - \gamma \left(\rvs^{(l)}+\epsilon \right)^{-\frac{1}{2}} \odot \rvg^{(l-1)}.
    \end{split}
\end{equation}
Here the square root, inverse, and $\odot$ are all element-wise operations.

\paragraph{RMSprop.} RMSprop~\citep{hinton2012neural} divides the gradient by a running average of its recent magnitude. To show that it is also a special case of~(\ref{eq:hidden_state}), let $\xi(\rvg) = \rvg^2$, $\varphi(\rvg) = \rvg$, and $\sigma(\rvs) = \left(\rvs^{(l)}+\epsilon \right)^{-\frac{1}{2}}$. Then the updating rule of RMSprop is 
\begin{equation}
    \begin{split}
        \rvs^{(l)} &= \beta \rvs^{(l-1)} + (1-\beta)(\rvg^{(l-1)})^2, \\
        \h{l} &= \h{l-1} - \left(\rvs^{(l)}+\epsilon \right)^{-\frac{1}{2}} \odot \rvg^{(l-1)}.
    \end{split}
\end{equation}

\paragraph{Adam.} Adam~\citep{kingma2014adam} requires two auxiliary variables $\rvs_1$ and $\rvs_2$. Let $\xi_1(\rvg) = \rvg$, $\xi_2(\rvg) = \rvg^2$, $\varphi(\rvg) = 1$, and $\sigma(\rvs^{(l)}_1, \rvs^{(l)}_2) = \left(\sqrt{\frac{\rvs^{(l)}_1}{1 - \beta_1^l}} + \epsilon\right)^{-1} \cdot \frac{\rvs^{(l)}_1}{1 - \beta_2^l}$. The updating rule of Adam is 
\begin{equation}
    \begin{split}
        \rvs_1^{(l)} &= \beta_1 \rvs_1^{(l-1)} - (1-\beta_1) \rvg^{(l-1)}, \\
        \rvs_2^{(l)} &= \beta_2 \rvs_2^{(l-1)} + (1-\beta_2) (\rvg^{(l-1)})^2, \\
        \h{l} &= \h{l-1} - \gamma \left(\sqrt{\frac{\rvs^{(l)}_1}{1 - \beta_1^l}} + \epsilon\right)^{-1} \cdot \frac{\rvs^{(l)}_1}{1 - \beta_2^l}.
    \end{split}
\end{equation}

\section{More on Label Propagation and Graph-Regularized MLPs} \label{sec: lp_gmlp}

In this section we pick up where we left off in Section \ref{sec:other_graph_ML}, filling in details at the nexus of label propagation, graph-regularized MLPs, and \model.


\subsection{Label Propagation as a \model Special Case} \label{sec:LP}
Consider the lower-level energy given by
\begin{equation}
\label{eq: energy_lp}
    \elow(\rmH; \gG) := \lambda \mbox{tr}\left(\rmH^\top \rmL \rmH\right) + \left\| \bar{\rmY} - \rmH \right\|_{\calF}^2 +  \sum_{v\in \calV'} \calI_\infty[\rvh_v \neq \bar{\rvy}_v],
\end{equation}
where $\rmL$ is the graph Laplacian of $\calG$ and $\bar{\rmY} \in \mathbb{R}^{n\times d}$ is constructed with $\bar{\rvy}_v = \rvy_v$ if $v \in \calV'$ and $\| \bar{\rvy}_v \| = 0$ otherwise.  This form is a special case of \model when we assume $m=1$, $f(\rvh_u,\rvh_v) = \lambda \|\rvh_u - \rvh_v \|_2^2$ with $\lambda > 0$ a hyperparameter, and $\rmX = \bar{\rmY}$, $\kappa(\rvh_v;\rvx_v) = \|\rvh_v - \rvx_v\|_2^2 \equiv \|\rvh_v - \bar{\rvy}_v\|_2^2$.  Furthermore, we generalize $\eta(\rvh_v) \rightarrow \eta(\rvh_v;\rvx_v) \equiv \eta(\rvh_v;\bar{\rvy}_v)$ and choose 
\begin{equation}
   \eta(\rvh_v;\bar{\rvy}_v) ~ = ~ \calI[\|\bar{\rvy}_v\| \neq 0 ] \calI_\infty[\rvh_v \neq \bar{\rvy}'_v],
\end{equation}
where we are implicitly assuming that $\| \bar{\rvy}_v \| \neq 0$ for any $v \in \calV'$.
The first two terms of (\ref{eq: energy_lp}) are smooth, while the last enforces that embeddings for nodes with observable labels must equal those labels.  We may minimize using proximal gradient steps of the form
 \begin{eqnarray} \label{eq:LP_update_general}
         \rmH^{(l)} & = & \mathbf{prox}_{\eta,\gamma}\left[ (1-\gamma) \rmH^{(l-1)} - \gamma \left(\lambda \rmL \rmH^{(l-1)} - \bar{\rmY} \right) \right] \nonumber \\
         & = & \calP_{\bar{Y}}\left[ (1-\gamma) \rmH^{(l-1)} - \gamma \left(\lambda \rmL \rmH^{(l-1)} - \bar{\rmY} \right) \right],
 \end{eqnarray}
where $\calP_{\bar{Y}}: \mathbb{R}^{n\times c} \rightarrow \mathbb{R}^{n\times c}$ is a projection operator that leaves rows $v \notin \calV'$ unchanged, while setting rows $v \in \calV'$ as $\rvh_v \rightarrow \bar{\rvy}_v = \rvy_v$.  Several points are worth noting regarding this result:
\begin{itemize}
    \item For different choices of $\gamma$ and $\lambda$, as well as the possible inclusion of a gradient preconditioner, the update from (\ref{eq:LP_update_general}) can exactly reproduce various forms of label propagation (LP) \citep{zhou2003learning,zhu2005semi}, which represents a family of semi-supervised learning algorithms designed to predict unlabeled nodes by propagating observed labels across edges of the graph.

\item As a representative LP example, we include a gradient preconditioner $\rmD^{-1}$ within (\ref{eq:LP_update_general}), where $\rmD$ is the diagonal degree matrix of $\calG$, and choose $\lambda \rightarrow \infty$, $\gamma \rightarrow 0$ such that $\lambda \gamma \rightarrow 1$.  Then (\ref{eq:LP_update_general}) reduces to the standard, interpretable form
\begin{equation}
 \rmH^{(l)}  =    \calP_{\bar{Y}}\left[ \rmD^{-1} \rmA \rmH^{(l-1)} \right],
\end{equation}
where $\rmA$ is the adjacency matrix of $\calG$.

    \item The input argument to $\calP_{\bar{Y}}[~\cdot~]$ in (\ref{eq:LP_update_general}) exactly reduces to (\ref{eq:simple_mp_case}) across all $v \in \calV$ if we replace the $\pi$ used in (\ref{eq:simple_mp_case}) with the corresponding labels $\bar{\rvy}_v$; this follows from the derivations presented in Section \ref{sec:derive_mp_special_case}. Moreover, the node-wise projection operator $\calP_{\bar{Y}}$ merely introduces a nonlinear activation function which can be merged into an update function $\hat{f}_U$.  In this way, executing the LP model propagation steps induces node embeddings analogous to those from a message-passing GNN architecture.

    \item If we include the $\eta$-based indicator term in (\ref{eq: energy_lp}), then w.l.o.g.~we can actually remove the $\kappa$-based term and still reproduce some notable LP variants.  However, we include the general form for two reasons: (i) It will facilitate more direct comparisons with graph-regularized MLP models below, and (ii) If we instead remove the $\eta$-based indicator term, then we can nonetheless recoup other flavors of LP without a projection step by retaining the $\kappa$-based term.

\item There exist more exotic LP models that are not directly covered by (\ref{eq: energy_lp}), such as ZooBP \citep{zoobp2017} and CAMLP \citep{camlp2016}, which include compatibility matrices to address graphs beyond vanilla homogeneous structure.  These can be accommodated by generalizing (\ref{eq: energy_lp}); however, we do not pursue this course further here.

\end{itemize}

We close by noting that, LP methods traditionally do not have parameters $\rmW$ to train, and the motivation for including an upper-level loss $\ell_{up}$ analogous to \model may be unclear.  However, recently trainable variants have been proposed \citep{wang2022propagate} that incorporate randomized sampling to mitigate label leakage.  \model can cover many such cases by supplementing (\ref{eq: energy_lp}) with parameters that can be subsequently trained using, for example, a node classification loss for $\ell_{up}$.

\subsection{Graph-Regularized MLPs, \model, and Connections with LP}
\label{sec: gmlp}

Graph-regularized MLPs (GR-MLP) represent a class of models typically applied to node classification tasks whereby no explicit message-passing occurs within actual layers of the architecture itself \citep{ando2006learning, hu2021graph, zhang2023orthoreg}.  Instead, a base model (like an MLP or related) is trained using a typical supervised learning loss combined with an additional regularization factor designed to push together the predictions of node labels sharing an edge.  These models are often motivated by their efficiency at inference time, where the graph structure is no longer utilized.

As a simple representative example,\footnote{Admittedly, this example does not cover the full diversity of possible GR-MLP models in the literature.  Nonetheless, it remains a useful starting point for analysis purposes.} consider the loss
\begin{equation}
\label{eq: energy_gmlp}
    \ell(\rmW; \gG) := \tfrac{1}{2}\|\bar{\rmY} - \rmH \|_{\calF}^2 + \tfrac{\lambda}{2} \mbox{tr}\left[\rmH^\top \rmL \rmH \right],~~\mbox{s.t.}~~\rvh_v = \pi(\rvx_v; \rmW)~~~ \forall v \in \calV.
\end{equation}
Here we are assuming that $\bar{\rmY}$ is defined analogously as in Section \ref{sec:LP}. In this expression, the first term represents quadratic supervision applied to node labels $\bar{\rmY}$, while the second term introduces the eponymous graph-regularization with graph Laplacian $\rmL$.  Within the constraint, $\pi$ is some node-wise trainable model such as an MLP.  Hence model training amounts to learning a $\rmW$ such that the resulting node embeddings closely match the given labels subject to smoothing across the graph. For this purpose, we take gradient steps
\begin{equation}\label{eq: gmlp_update}
    \rmW^{(l)} = \rmW^{(l-1)} - \gamma \left. \frac{\partial \ell(\rmW; \gG)}{\partial  \rmW} \right|_{\rmW = \rmW^{(l-1)}}, ~~~ l = 1,\ldots, L.
\end{equation}
We now show how these steps induce implicit message-passing GNN-like structure in certain cases, and later, connect to label propagation.  To see this effect in transparent terms, assume that $\pi(\rvx;\rmW) = \rvx \rmW$.  Then we have
\begin{equation}
\frac{\partial \ell(\rmW ; \gG)}{\partial  \rmW} = \rmX^\top \left(\rmH + \lambda \rmL \rmH - \bar{\rmY} \right).
\end{equation}
From here we define $\rmH^{(l)} := \rmX \rmW^{(l)}$ such that the above gradient steps w.r.t.~$\rmW$ produce a corresponding sequence of embeddings given by
\begin{equation} \label{eq:H_update_graph_MLP}
    \rmH^{(l)} = \rmH^{(l-1)} - \gamma \rmX \rmX^\top \left(\rmH^{(l-1)} + \lambda \rmL \rmH^{(l-1)} - \bar{\rmY} \right).
\end{equation}
Provided that $\rmX$ is full column rank, we can assume w.l.o.g.~that columns of $\rmX$ are orthonormal, i.e., $\rmX^\top \rmX = \rmI$. (This is possible because we can always convert non-orthonormal columns into orthonormal ones via an invertible linear transformation that can be absorbed into $\rmW$.) Consequently, we may directly conclude that $\rmX \rmX^\top = \calP_X$, where $\calP_X$ indicates an orthogonal projection to the range of $\rmX$, denoted $\mbox{range}[\rmX]$.  We then provide the following straightforward interpretations of this process:
\begin{itemize}

\item  From the perspective of \model, we can convert (\ref{eq: energy_gmlp}) to an equivalent $\ell_{low}$ given by
\begin{equation} \label{eq:L_low_for_GR_MLP}
\ell_{low}(\rmH;\rmW,\gG) := \tfrac{\lambda}{2} \sum_{(u,v) \in \gE} \left\|\rvh_u - \rvh_v \right\|_{2}^2 + \sum_{v \in \gV} \tfrac{1}{2}\left\|\rvh_v - \bar{\rvy}_v  \right\|_{2}^2 +  \gI_{\infty}\{ \rmH \notin \mbox{range}[\rmX] \},
\end{equation}
where we are treating the labels from $\bar{\rmY}$ as auxiliary inputs to the $\kappa$ function, and we are trivially generalizing the non-smooth $\eta$ function to apply across all of $\rmH$ while including a dependency on the original node features from $\rmX$.  Additionally, the proximal gradient update for minimizing (\ref{eq:L_low_for_GR_MLP}) is given by
\begin{equation} \label{eq:H_update_graph_MLP_prox_vers}
    \rmH^{(l)} = \calP_X\left[ (1-\gamma) \rmH^{(l-1)} - \gamma \left(\lambda \rmL \rmH^{(l-1)} - \bar{\rmY} \right) \right],
\end{equation}
which is equivalent to (\ref{eq:H_update_graph_MLP}) provided $\rmH^{(0)} \in \mbox{range}[\rmX]$.  Analogous to the LP derivations from Section \ref{sec:LP}, the input argument to $\calP_X[~\cdot~]$ in (\ref{eq:H_update_graph_MLP_prox_vers}) exactly reduces to (\ref{eq:simple_mp_case}) across all $v \in \calV$ if we replace the $\pi$ used in (\ref{eq:simple_mp_case}) with the corresponding labels $\bar{\rvy}_v$; this follows from the derivations presented in Section \ref{sec:derive_mp_special_case}.  In this way, training GR-MLP model weights induces node embeddings analogous to those from a message-passing GNN architecture, the final projection operator notwithstanding.

\item If $d=n$, then $\calP_X = \rmX \rmX^\top = \rmI$ and we are operating in an overparameterized regime where $\rmH = \rmW$.  In this special case, the iterations from (\ref{eq:H_update_graph_MLP_prox_vers}) exactly reproduce LP from Section \ref{sec:LP}, excluding the projection step as some LP variants do.

    \item In the more typical setting with $d< n$, the lingering difference between (\ref{eq:H_update_graph_MLP_prox_vers}) and LP hinges on the projection operator being invoked, $\calP_X$ versus $\calP_{\bar{Y}}$ (where $\calP_{\bar{Y}}$ is defined in Section \ref{sec:LP}; it is \textit{not} equivalent to a projection to $\mbox{range}[\bar{\rmY}]$).  Which operator is to be preferred depends heavily on the quality of input features.  If $\mbox{range}[\rmX]$ closely aligns with $\mbox{range}[\rmY]$ (the range space of the true label matrix), then $\calP_X$ will be preferable.  In contrast, when the relationship is closer to $\mbox{range}[\rmX] \perp \mbox{range}[\rmY]$, then $\calP_{\bar{Y}}$ will generally be superior.
\end{itemize}

\subsection{Empirical Illustration of LP/GR-MLP Insights}
To illustrate some of the concepts from the previous section, we conduct a simple experiment involving Cora and Citeseer data.  We choose $\pi(\rvx; \rmW) = \rvx \rmW$ and train GR-MLP models based on (\ref{eq:H_update_graph_MLP_prox_vers}) using (i) original node features, and (ii) overparameterized features whereby $d=n$.  We also compare with an LP model, excluding the projection step for a more direct evaluation.  

Node classification results from these experiments are shown in Table \ref{tab: lp_gmlp}.  The first two rows confirm the equivalence between GR-MLP and LP for the reasons detailed in Section \ref{sec: gmlp}.  Furthermore, by examining the second and third rows, we are able to observe how the quality of node features influences the efficacy of $\calP_X$.  Specifically, it has been previously demonstrated that the features and labels of Cora nodes have minimal positive (linear) correlation \citep{luo2021reslpa}.  Therefore, from Table \ref{tab: lp_gmlp} we see that the GR-MLP model using the original features performs much worse than the overparameterized version.  In contrast, with Citeseer where the input features are more aligned with the labels, the original feature model significantly outperforms the overparameterized one.

\begin{table}[h]
\caption{\textit{Node classification accuracy (\%) comparisons}. Results are averaged over 5 trials; error bars (not shown) are negligible.}
\label{tab: lp_gmlp}
\centering
\begin{tabular}{l|cc}
\specialrule{.15em}{.05em}{.05em}
                               & Cora & Citeseer          \\ \hline
Label Propagation              & 70.2 & 50.2       \\
GR-MLP Overparam. & 70.2 & 50.2    \\
GR-MLP Original         & 60.4 & 64.1     \\
\specialrule{.15em}{.05em}{.05em}
\end{tabular}
\end{table}

\section{Model Implementation and Section \ref{sec:experiments} Experiment Details} \label{sec: exper_details}

In this section we provide further details regarding the experiments from Section \ref{sec:experiments}. 

\paragraph{Details for Table~\ref{tab: noise}.}

 Our implementation is based on modifications of the public codebase  from~\cite{yang2021graph}. We adopt the hyperparameters they reported for all datasets. For \model, we apply a vectorized version of a Huber penalty in the form 
\begin{equation} \label{eq:huber_basic_definition}
    \delta(\rvu ) := \sum_{i=1}^d \left\{
    \begin{array}{ll}
         \frac{1}{2} u_i^2, &  \mbox{for}~~|u_i| < 1 \\
          |u_i| - \frac{1}{2}, &  \mbox{otherwise},
    \end{array}
    \right.
\end{equation}
where $\{u_i\}_{i=1}^d$ are the elements of the input vector $\rvu \in \mathbb{R}^d$. We then form the energy function 
\begin{equation}\label{eq:huber_energy}
    \ell_{low}(\rmH; \rmW,\calG) = \sum_{(u, v)\in \gE} f(\rvh_u, \rvh_v;r) + \sum_{v \in \gV} \kappa(\rvh_v; \rvx_v) = \frac{\lambda}{2} \sum_{(u, v)\in \gE} \left\|\rvh_u - \rvh_v\right\|^2_2 + \sum_{v \in \gV} \delta\left(\rvh_v - \pi(\rvx_v;\rmW) \right),
\end{equation}
where the dependency of $f$ on $r$ is irrelevant for homogeneous graphs.  Note that because (\ref{eq:huber_energy}) is smooth and differentiable (i.e., the Huber loss is differentiable and the $\ell_1$ norm defaults to merely a summation since its argument is non-negative), no $\eta$ term is needed.  Hence, we choose $\calA$ as gradient descent for producing the \model model $\calA[\ell_{low},~\cdot ~]$.  For $\ell_{up}$ we apply a standard node classification loss.  We reproduce Table~\ref{tab: noise} from the main text with error bars obtained from averaging over 10 trials;  these results are presented in~Table~\ref{tab: noise_full}.

\begin{table}[h]
\captionsetup{font=small}
\small
\caption{Full version of Table~\ref{tab: noise} including error bars from averaging over 10 runs. \textit{Performance mitigating spurious input features.} Here the detect ratio is obtained by computing $\|\rvh_v - \pi(\rvx_v;\rmW)  \|_2$ for all $v\in \calV$ and then segmenting out the percentage of corrupted nodes within the largest 20\%.}
\label{tab: noise_full}
\centering
\begin{tabular}{cc|cccc}
\specialrule{.15em}{.05em}{.05em}
&   & Cora  & Citeseer & Pubmed & Arxiv \\ \hline
\multirow{2}{*}{Accuracy}& Base               & 54.97 $\pm$ 1.95 &   38.98 $\pm$ 1.96    & 45.67 $\pm$ 3.21  &     47.63 $\pm$ 0.23   \\
& \model    & \textbf{65.83 $\pm$ 2.28} &  \textbf{49.33 $\pm$ 1.95}    & \textbf{72.70 $\pm$ 1.58}  & \textbf{69.11 $\pm$ 0.23}   \\ \hline
Detect Ratio & \model & 94.27 & 87.82 & 96.98 & 100.00\\
\specialrule{.15em}{.05em}{.05em}
\end{tabular}
\end{table}

\paragraph{Details for Table~\ref{tab: heter}.}

Building on the results from above and the description in Section \ref{sec:experiments} in the main text, we consider the generalized form
\begin{equation}\label{eq:huber_energy2}
    \ell_{low}(\rmH; \rmW,\calG) = \sum_{(u, v)\in \gE} f(\rvh_u, \rvh_v;r) + \sum_{v \in \gV} \kappa(\rvh_v; \rvx_v) = \frac{\lambda}{2} \sum_{(u, v)\in \gE} \left\|\rvh_u \rmC - \rvh_v\right\|^2_2 + \sum_{v \in \gV}  \delta\left(\rvh_v - \pi(\rvx_v;\rmW) \right).
\end{equation}
As before, this expression is differentiable so we can choose $\calA$ as gradient descent with step-size parameter $\gamma$. All of the hyperparameters were selected via Bayesian optimization using \textit{Wandb}\footnote{https://wandb.ai/site} and we also provide the model hyperparameters search space in~Table~\ref{tab: bo_hyper}.  Note that we also found that it can be effective to simply learn $\lambda$, i.e., absorb it into $\rmW$, which reduces the hyperparameter search space and can potentially even improve performance.  We show such supporting ablations in Section \ref{sec:learning_lambda}.

In Table~\ref{tab: heter_full} we reproduce results from Table~\ref{tab: heter}, including error bars from averaging over 10 runs and an additional ablation.  For the latter, we compare $\kappa(\rvh;\rvx) = \| \rvh - \pi(\rvx;\rmW)\|_2^2$ labeled as \model (w/o Huber) with the $\kappa(\rvh;\rvx) = \delta\left(\rvh_v - \pi(\rvx_v;\rmW) \right) $ implied within (\ref{eq:huber_energy2}). From Table~\ref{tab: heter_full} we observe that the Huber version works better on 4 of 5 datasets, the exception being Amazon where it is significantly worse.  As one candidate explanation for this exception, we note that the Amazon dataset has a significantly higher \textit{adjusted homophily} ratio than the other 4 datasets, where adjusted homophily is defined as in \cite{platonov2022characterizing} (this metric is less sensitive to the number of classes and their balance).  

We also trained several standard GNN architectures, e.g., GCN and GAT, for reference purposes; however, we found that their performance was considerably worse than the heterophily baseline architectures from Table~\ref{tab: heter_full} (and therefore Table~\ref{tab: heter} as well).  This is in contrast to \cite{platonov2023a}, which presents somewhat stronger results associated with such popular architectures.  However, this apparent contradiction can be resolved by closer examination of the public implementations from \cite{platonov2023a}.  Here we observe that additional modifications to the original models have been introduced to the codebase that can significantly alter performance.  While certainly valuable to consider, these types of enhancements can be widely applied and lie beyond the scope of our bilevel optimization lens.  Indeed the flexibility of \model also readily accommodates further architectural adjustments with the potential to analogously improve performance as well on specific heterophily tasks.



\begin{table}[h]
\centering
\captionsetup{font=small}
\small
\caption{Full version of Table~\ref{tab: heter} including additional ablations and error bars from averaging over 10 runs. \textit{Node classification on heterophily graphs}.
Following convention, accuracy (\%) is reported for Roman and Amazon, while ROC-AUC (\%) is used for Minesweeper, Tolokers, and Questions. The numbers in the upper block are from~\cite{platonov2023a}.}
\label{tab: heter_full}
\begin{tabular}{l|ccccc|c}
\specialrule{.15em}{.05em}{.05em}
& Roman     & Amazon   & Minesweep      & Tolokers         & Questions & Avg.        \\ \hline
FAGCN   &65.22 $\pm$ 0.56 & 44.12 $\pm$ 0.30 & 88.17 $\pm$  0.73 & 77.75 $\pm$ 1.05 & 77.24 $\pm$ 1.26 & 70.50 \\
FSGNN      & 79.92 $\pm$ 0.56 & 52.74 $\pm$ 0.83  & 90.08 $\pm$  0.70  &  82.76 $\pm$ 0.61  & \textbf{78.86 $\pm$ 0.92} & 76.87\\
GBK-GNN       & 74.57 $\pm$ 0.47   & 45.98 $\pm$ 0.71 & 90.85 $\pm$ 0.58  & 81.01 $\pm$ 0.67  & 74.47 $\pm$ 0.86  & 73.38\\ 
JacobiConv      & 71.14 $\pm$ 0.42 &  43.55 $\pm$ 0.48  & 89.66 $\pm$ 0.40  &  68.66 $\pm$ 0.65 & 73.88 $\pm$ 1.16 & 69.38\\ \hline
Base from (\ref{eq:quad_example_energy})      & 76.63 $\pm$ 0.24 & 52.37 $\pm$ 0.20 & 88.97 $\pm$ 0.05 & 80.91 $\pm$ 0.24 & 76.72 $\pm$ 0.49 & 75.12  \\
\model (w/o Huber) & 84.45 $\pm$ 0.31 & \textbf{52.92 $\pm$ 0.39} & 91.83 $\pm$ 0.28 & 84.84 $\pm$ 0.27 & 77.62 $\pm$ 0.33 & 78.33 \\
\model (w/ Huber) & \textbf{85.26 $\pm$ 0.25} & 51.00 $\pm$ 0.47 & \textbf{93.30 $\pm$ 0.16} & \textbf{85.92 $\pm$ 0.14} & 77.93 $\pm$ 0.34 & \textbf{78.68} \\
\specialrule{.15em}{.05em}{.05em}
\end{tabular}
\end{table}

\begin{table}[h]
\centering
\caption{Model hyperparameters selection search space for Tables~\ref{tab: heter} and~\ref{tab: halo}. For \model (w/o Huber), $\lambda$ is a learnable parameter that is not searched by BO; see Section \ref{sec:learning_lambda} for an ablation on learning $\lambda$.}
\label{tab: bo_hyper}
\begin{tabular}{l|c|c}
\specialrule{.15em}{.05em}{.05em}
   Hyperparameters   &     Range for Table~\ref{tab: heter}   & Range for Table~\ref{tab: halo}     \\ \hline
$L$ &  [2,4,6]  & [4, 8, 16]    \\
$\lambda$    & [0.1 , 1, 5, 10]    & [0.01, 0.1 , 1, 10]          \\
$\gamma$   & [0 , 0.5, 1]    & [0 , 0.1, 0.5, 1]          \\
MLP layers before prop  & [0, 1, 2]     & [0, 1, 2]        \\
Hidden size  & [128,256,512]  & [16, 64, 128]         \\
\specialrule{.15em}{.05em}{.05em}
\end{tabular}
\end{table}

\paragraph{Details for Figure~\ref{fig:norm_dist}.}
To show the effect of \model to have connected nodes with deviated embeddings, we use a Huber penalty for the edge-dependent regularizer term leading to the energy function 
\begin{equation}
    \ell_{low}(\rmH; \rmW,\calG) = \sum_{(u, v)\in \gE} f(\rvh_u, \rvh_v;r) + \sum_{v \in \gV} \kappa(\rvh_v; \rvx_v) = \frac{\lambda}{2} \sum_{(u, v)\in \gE} \delta(\rvh_u - \rvh_v) + \frac{1}{2} \sum_{v \in \gV} \left\|\rvh_v - \pi(\rvx_v; \rmW)  \right\|_{2}^2. 
\end{equation}
For the experiments, we fixed the hidden embedding size to 1024, which gives the model more capacity for handling heterphility, and we run Bayesian optimization with the same range as in the left block of~Table~\ref{tab: bo_hyper} for the remaining hyperparameters.

\paragraph{Details for Table~\ref{tab: halo}.}  For these results involving heterogeneous graphs, we modify the codebase from the HALO model~\citep{ahn2022descent} and adopt $\kappa(\rvh;\rvx) = \delta(\rvh_v - \pi(\rvx_v;\rmW))$ or $\kappa(\rvh;\rvx) = \omega(\rvh_v - \pi(\rvx_v;\rmW))$, where $\omega$ is the Log-Cosh loss given by
\begin{equation} \label{eq:logcosh_basic_definition}
    \omega(\rvu ) := \sum_{i=1}^d \log(\cosh{u_i}).
\end{equation}
The hyperparameters we used are also selected via Bayesian optimization with Wandb; the search range is provided in the right block of Table~\ref{tab: bo_hyper}, for both HALO and \model. Note that we observe negligible variance in the test accuracies across different seeds, so we omit error bars from multiple runs.

\paragraph{Details for Table~\ref{tab: kgc}.}
We provide specifics directly tied to creating Table~\ref{tab: kgc} here, while a more comprehensive treatment of how NBFNet relates to \model is deferred to Section~\ref{sec: nbfnet}. Our implementation is based on modifications of the public codebase from~\cite{zhu2021neural}, which involves (among other things) two trainable linear transformations. The first involves parameters $\rmW_r$ and $\rvb_r$ used to create the relation embeddings for each relation type via $\rve_r = \rmW_r \rvq_{q} + \rvb_r$, where $\rvq_{q}$ is the query embedding. The other is $\Phi$ used to transform the output of the aggregation function back to the embedding space. To satisfy the criteria of \model, we modify the code to share these weights across all layers. 
Performance results for RefactorGNN~\citep{chen2022refactor} were taken from the original paper due to lack of publicly-available reproducible code for conducting our own comparisons. 

\paragraph{Details for Table~\ref{tab: momentum} and Figure~\ref{fig: energy}.}

For Table~\ref{tab: momentum}, we modify the codebase from~\cite{yang2021graph} by using the update step defined in (\ref{eq:momentum_update_fuction}). The components are chosen based on (\ref{eq:simple_mp_case}). The hyperparameters are chosen via grid search over $\lambda$, $\gamma$ and $\beta$. For Figure~\ref{fig: energy}, we use the same set of hyperparameters for both SGD and \model (momentum), with the number of message-passing layers being 200.


\section{Exploring the Connection between NBFNet and \model}
\label{sec: nbfnet}

Given a graph $\calG = (\calV,\calR, \calE)$, let $\gG_{cq}$ denote a conditional version with the same node, relation, and edge set, but with node features $\rmX_{cq}= \{\rvx_{v, cq} \}_{v \in \calV}$ and labels $\rmY_{cq} = \{\rvy_{v, cq} \}_{v \in \calV}$ that depend on a fixed source node $c \in \calV$ (in this section $c$ should not be confused with the class label dimension used elsewhere) and query relation $q \in \calR$.  By conditioning in this way, KGC or link prediction queries specific to $c$ and $q$ can be converted to node classification, e.g. $\rvy_{v,cq} = 1$ if $(c,q,v)$ is a true edge in the original graph.  We may then later expand to unrestricted link prediction tasks by replicating across all $c \in \calV$ and $q \in \calR$.  This is the high-level strategy of NBFNet.

In this section, we first show that given a graph $\gG_{cq}$ described as above, an NBFNet-like architecture with parameters shared across layers can be induced using \model iterations of the form $\calA[\ell_{low},~\cdot~]$ for suitable choice of energy $\ell_{low}$ and $\calA$.  For this purpose, we define the lower-level energy function as 
\begin{eqnarray}\label{eq: energy_nbf_phi}
&& \hspace*{-0.8cm} \elow(\rmH_{cq}; \rmW, \gG_{cq})  :=  \\
&& \sum_{(u,r,v) \in \gE_{cq}} \left(\rvh_{v,cq}^\top\Phi\rvh_{u,cq}+ \rvh_{v,cq}^\top\Phi\rve_r + \rvh_{u,cq}^\top \Phi\rve_{r^{-1}} \right)+ \frac{1}{2} \sum_{v \in \gV_{cq}} \left[|| \Psi^{\frac{1}{2}} \rvh_{v,cq} + \Psi^{-\frac{1}{2}}\Phi \rvx_{v,cq} ||^2_2 + \eta(\rvh_{v,cq}) \right], \nonumber 
\end{eqnarray}
where $\rmH_{cq} = \{\rvh_{v, cq} \}_{v \in \calV}$ are node embeddings and $\rve_r := \rmW_r \rvq + \rvb_r$ are relation embeddings, $\rvq$ is a so-called shared query relation embedding (distinct from the relation index of query $q$), and $\rmW_r$ and $\rvb_r$ are trainable parameters that are bundled within $\rmW$ along with $\Phi, \Psi \in \sR^{d \times d}$.  Additionally, $\rvx_{v, cq} := \gI(c = v) \rvq$  for all $v \in \gV_{cq}$. To ensure the gradient expressions are symmetric in form across relations and inverse relations, we require that $\Phi$ and $\Psi$ are symmetric matrices.  However, it has been shown in~\cite{yang2021implicit} that in certain circumstances, asymmetric weights can be reproduced by symmetric ones via an appropriate expansion of the hidden dimensions.
Let 

\begin{equation}
    f(\rvh_{u,cq}, \rvh_{v,cq}; r) = \rvh_{v,cq}^\top\Phi\rvh_{u,cq}+ \rvh_{v,cq}^\top\Phi\rve_r + \rvh_{u,cq}^\top \Phi\rve_{r^{-1}}
\end{equation}

Then we have

\begin{equation}
\begin{aligned}
    \frac{\partial f\left(\rvh_{u,cq}, \rvh_{v,cq}; r \right)}{\partial \rvh_{v,cq}} &=  \Phi \rvh_{u,cq} + \Phi \rve_{r} =: \Phi \widetilde{f}_M(\rvh_{u,cq}, r, \rvh_{v,cq})\\
    \frac{\partial f\left(\rvh_{u,cq}, \rvh_{v,cq}; r \right)}{\partial \rvh_{u,cq}} &=  \Phi \rvh_{v,cq} + \Phi \rve_{r^{-1}} =: \Phi \widetilde{f}_M(\rvh_{u,cq}, r^{-1}, \rvh_{v,cq})
\end{aligned}
  \end{equation}

Thus, the aggregated function is 

\begin{equation}
    \rva_{v,cq} = \sum_{(u,r) \in \gN_v}\widetilde{f}_M(\rvh_{u,cq}, r, \rvh_{v,cq}) = \sum_{(u,r) \in \gN_{v,cq}} \left(\rvh_{u,cq} + \rve_{r}\right)
\end{equation}


Next, let 
\begin{equation}
    \kappa(\rvh_{v,cq}; \rvx_{v,cq}) = || \Psi^{\frac{1}{2}} \rvh_{v,cq} + \Psi^{-\frac{1}{2}}\Phi \rvx_{v,cq} ||^2_2
\end{equation}

Then, by reintroducing $\eta$ and adopting proximal gradient descent for $\calA$, we produce the message-passing function
\begin{equation}\label{eq:update_nbf_phi}
\begin{aligned}
    & \vmu_{(u, r, v), cq}^{(l)} := \rvh_{u,cq}^{(l-1)} + \rve_r^{(l-1)} \\
    & \rva_{v, cq}^{(l)} := \sum_{(u, r) \in \gN_{v, cq}}  \vmu_{(u, r, v), cq}^{(l)}  \\
    & \rvh_{v,cq}^{(l)} := \mathbf{prox}_{1, \eta}\left(\Phi \left[\rva_{v, cq}^{(l)} + \rvx_{v, cq} \right] + \Psi \rvh_{v, cq}^{(l-1)}  \right).
\end{aligned}
\end{equation}
This expression closely resembles an NBFNet model layer with summation aggregation.

Proceeding further, we extend the scope of our discussion to an expanded graph $\gG_{exp}$ with node set $\gV_{exp}$ and edge set $\gE_{exp}$, where $\gG_{exp} = \cup \gG_{cq},~\forall (c, q, \cdot) \in \gE_{exp}$. Analogously, we define an extended energy function over $\gG_{exp}$ as
\begin{equation}
\begin{aligned}
    \elow(\rmH; \rmW, \gG_{exp}) &:= \sum_{(c, q, \cdot) \in \gE_{exp}} \elow(\rmH_{cq}; \rmW, \gG_{cq}) \\
    &\hspace*{-2cm} = \sum_{(c, q, \cdot) \in \gE_{exp}}\Bigg( \Bigg.\sum_{(u,r,v) \in \gE_{cq}} \left(\rvh_{v,cq}^\top\Phi\rvh_{u,cq}+ \rvh_{v,cq}^\top\Phi\rve_r + \rvh_{u,cq}^\top \Phi\rve_{r^{-1}} \right)\\ &+ \frac{1}{2} \sum_{v \in \gV_{cq}} \left[|| \Psi^{\frac{1}{2}} \rvh_{v,cq} + \Psi^{-\frac{1}{2}}\Phi \rvx_{v,cq} ||^2_2 + \eta(\rvh_{v,cq}) \right] \Bigg) \Bigg.
\end{aligned}
\end{equation}
For elements in $\rmY_{cq}$, positive instances correspond with true edges in the original graph, while negative instances are obtained by sampling.   With $\rmY_{exp} := \cup \rmY_{cq}$, we define an upper-level energy as
\begin{equation}
    \ell_{up}(\rmW, \rmY_{exp}; \Theta) := \sum_{(c, q, \cdot) \in \gE_{exp}} \sum_{v \in \gV'_{cp}} \gD\left[ g \left(\rvh^{(L)}_{v, cq}(\rmW) ; \Theta \right), \rvy_{v,cq}\right],
\end{equation}
where $\gV'_{cq}$ represents the node set of each $\gG_{cq}$ with observed labels.  In this way, the resulting bilevel optimization process can be viewed as instantiating a \model node classification task on the expanded graph $\calG_{exp}$, with architecture mimicking a specialized NBFNet-like model designed to minimize $\ell_{low}$.

\section{Additional Experiments} \label{sec:additional_experiments}


In this section we include additional ablations and further demonstration of the versatility of \model.

\subsection{Ablation on Learning $\lambda$} \label{sec:learning_lambda}

There are two potential advantages to learning differentiable hyperparameters within $\ell_{low}$.  First, relative to hyperparameter tuning, it may be possible to boost performance.  And secondly, even if performance improvements are stubborn, by learning such hyperparameters, we economize the sweep over any remaining hyperparameters as the effective search space has compressed along one dimension.  In this regard, Table \ref{tab:heter_ablation} shows results on the heterophily datasets (and the same setup from Tables \ref{tab: heter} and \ref{tab: heter_full}) with and without learning $\lambda$.  From these results we observe that learning $\lambda$ produces reliable performance, suggesting that it can be trained instead of tuned (as a hyperparameter) without sacrificing accuracy.

\begin{table}[h]
\centering
\captionsetup{font=small}
\small
\caption{\textit{Ablation on learnable} $\lambda$.  Accuracy results are obtained by averaging over 10 trials.}
\label{tab:heter_ablation}
\begin{tabular}{l|ccccc|c}
\specialrule{.15em}{.05em}{.05em}
& Roman     & Amazon   & Minesweep      & Tolokers         & Questions & Avg.        \\ \hline
w/o Huber      &     84.28 $\pm$ 0.63 & 52.76 $\pm$ 0.27 & 91.89 $\pm$ 0.25 & 84.66 $\pm$ 0.33 & 77.46 $\pm$ 0.30 & 78.21 \\
w/o Huber (learnable $\lambda$) & 84.45 $\pm$ 0.31 & \textbf{52.92 $\pm$ 0.39} & 91.83 $\pm$ 0.28 & 84.84 $\pm$ 0.27 & 77.62 $\pm$ 0.33 & 78.33 \\
w/ Huber & 85.26 $\pm$ 0.25 & 51.00 $\pm$ 0.47 & \textbf{93.30 $\pm$ 0.16} & \textbf{85.92 $\pm$ 0.14} & 77.93 $\pm$ 0.34 & \textbf{78.68} \\
w/ Huber (learnable $\lambda$) & \textbf{85.36 $\pm$ 0.36} & 51.32 $\pm$ 0.43 & 92.70 $\pm$ 0.60 & 85.86 $\pm$ 0.18 & \textbf{78.01 $\pm$ 0.45} & 78.65 \\
\specialrule{.15em}{.05em}{.05em}
\end{tabular}
\end{table}

\subsection{Efficiently Enforcing Layernorm Using $\eta$}

\begin{figure}[h]
 \captionsetup{font=footnotesize}
  \centering
  \includegraphics[width=0.4\textwidth]{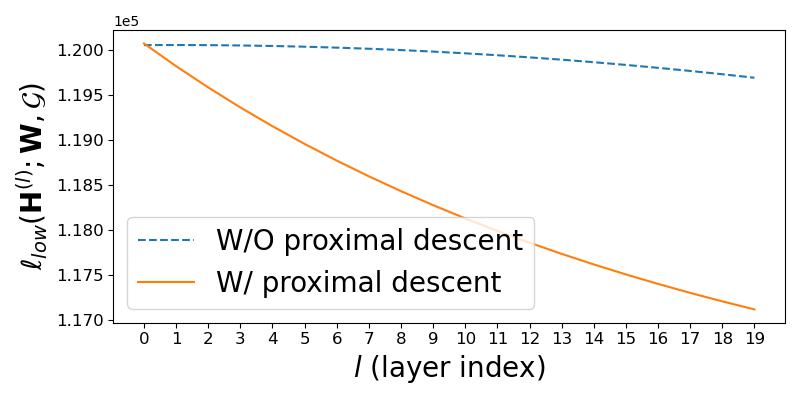} 
   \vspace{-0.3cm}
  \caption{$\elow$ value versus the number of propagation steps in Roman dataset.}
  \label{fig: layernorm}
\end{figure}

If for some reason we prefer to have a model with layer-normalized node embeddings, one option is to introduce a penalty of the form $(\|\rvh\|_2^2 - 1)^2$, absorb this factor into $\kappa(\rvh;\rvx)$, and pthen roceed to minimize $\ell_{low}$ using regular gradient descent for $\calA$.  However, this requires an additional trade-off hyperparameter, and if we want to be arbitrarily close to the unit norm, convergence will become prohibitively slow.  Alternatively, we can enforce strict adherence to unit norm by choosing $\eta(\rvh) = \gI_{\infty}\{\|\rvh\|_{2}^{2} \neq 1\}$ and then adopt proximal gradient descent for $\calA$ to handle the resulting discontinuous loss surface.  This involves simply taking gradient steps on all other differentiable terms and then projecting to the unit sphere (i.e., the proximal operator).  We compare these two approaches in Figure \ref{fig: layernorm}, where proximal gradient descent converges far more rapidly. For the model without proximal gradient descent, we add the penalty $\alpha (\|\rvh\|_2^2 - 1)^2$ to the energy from~(\ref{eq:quad_example_energy}), where $\alpha$ is the trade-off parameter; we choose $\alpha=1$ in Figure~\ref{fig: layernorm}. For the model with proximal gradient descent, we project each node embedding to the surface of a ball satisfying $||\rvh||_2^2 = 1$. 

\end{document}


\appendix

\section*{Appendix}

\section{Details of Experiments}

\subsection{Set-up of Table~1 and Figure~1}

We use the code base from TWIRLS \cite{yang2021graph} and change the propagation step based on the loss we proposed. Specifically, we did grid search and ran each of the experiments for 5 rounds (including the baseline). 

For Figure~1, we set the unfolding step size to 0.1. We show the energy function values along 200 unfolding layers on test set.

\subsection{Set-up of Table~2}

For label propagation, we use a linear layer as the model, and set the input feature as an identity matrix, which is equivalent to training an embedding layer sharing the same shape with the linear layer parameters. For GR-MLP with orthogonal features, we pad the columns of input $X$ and implement SVD to get the orthogonal matrix as the input.

\subsection{Set-up of Table~3}

We use the code base of HALO \cite{ahn2022descent} and modify the unfolding propagation class based on the gradient of Log-Cosh and Huber loss.

\subsection{Discussion of NBFNet as a Special Case}

Here we show that NBFNet \cite{zhu2021neural} as adopted to obtain results in Table~4 is a special case of the general bilevel optimization framework we have proposed. NBFNet is a graph learning framework for obtaining node representations conditioned on the source node. The propagating step is given by

\begin{equation}
    \begin{split}
        &h_0[v] \leftarrow \mathrm{Indicator}(u, r, v) \\
        &h_l[v] \leftarrow f_A\left( \left\{f_M\left( h_{l-1}[u], r, h_{l-1}[v] \right): (u, r) \in \mathcal{N}^1[v] \right\} \cup \left\{ h_0[v] \right\}\right),
    \end{split}
\end{equation}

where $f_A$ and $f_M$ are aggregation and message functions which follow the rule of message-passing propagation. In \cite{zhu2021neural}, $f_A(\cdot)$ is a summation function over $\mathcal{N}^1[v]$ and the initial self-embedding. Thus, if we use the same notation as in Definition~2 and let $f_U(h_{l-1}[v], a_l[v]) = a_l[v]$, then the corresponding energy function is 
\begin{equation}
    \mathcal{L}(h, \mathcal{T}) = \sum_l \mathcal{L}(h_{l}, \mathcal{T}) = \sum_l \sum_{(u, r, v) \in \mathcal{T}} f\left((h_l[u], r, h_l[v])\right)
\end{equation}
where $\nabla_{h_{l-1}[v]}f\left((h_{l-1}[u], r, h_{l-1}[v])\right) = a_l[v]$. With learning rate $\alpha = 1$, we have the unfolding gradient step on node $v$ in $l$-th layer given by 

\begin{equation}
    h_l[v] = (1 - \alpha) h_{l-1}[v] + \alpha \nabla_{h_{l-1}[v]} \mathcal{L}(h_{l-1}, \mathcal{T}).
\end{equation}

After obtaining the learned embeddings $h$, as illustrated in \cite{zhu2021neural}, they take it into a likelihood function for the higher-level optimization.

\section{Proof of Theorem~1}

Recall that the energy we proposed is \begin{equation}\label{eq: energy}
            \mathcal{L}(h, \mathcal{T}) = \sum_l \mathcal{L}(h_{l}, \mathcal{T}) = \sum_l \sum_{(u, r, v) \in \mathcal{T}} f\left((h_l[u], r, h_l[v])\right).
        \end{equation}

\paragraph{From (i) to (ii)} we prove that for any $\mathcal{L}$ satisfying Equation~\ref{eq: energy}, its unfolding gradient step is a message-passing layer and its gradient is permutation invariant.

Here we show how to define the three functions to enable the unfolding gradient step work as a message-passing layer. Taking the Eq~\ref{eq: energy} into the gradient operator, we have

\begin{equation}
    \begin{split}
        h_l[v] = \mathbf{GD}(h_{l-1}, \mathcal{T})[v] &= h_{l-1}[v] - \alpha \nabla_{h_{l-1}[v]} \mathcal{L}(h_{l-1}, \mathcal{T})  \\
        &= h_{l-1}[v] - \alpha  \sum_{ (u, r) \in \mathcal{N}^1[v]}\left[\nabla_{h_{l-1}[v]}f\left((h_{l-1}[u], r, h_{l-1}[v])\right)\right].
    \end{split}
\end{equation}

Then, for a specific destination node $v$, let $m[u, r, v] = \nabla_{h[v]} f\left((h[u], r, h[v])\right)$, and we have

\begin{equation}
    h_l[v] = h_{l-1}[v] - \alpha \sum_{ (u, r) \in \mathcal{N}^1[v]}m_l[u, r, v],
\end{equation}

where $$a_l[v] = f_A(\{m_l[u, r, v]: (u, r) \in \mathcal{N}^1[v]\}) = \sum_{ (u, r) \in \mathcal{N}^1[v]}m_l[u, r, v]$$ and $$f_U(h_{l-1}[v], a_l[v]) = h_{l-1}[v] - \alpha a_l[v].$$

\textbf{From (ii) to (i)} we prove that any energy function whose unfolding gradient step can be expressed as a message-passing layer should be in the form of Equation~$\ref{eq: energy}$.

Recall that we have

$$
h_l = \mathbf{GD}(h_{l-1}, \mathcal{T}) = h_{l-1} - \alpha \nabla_{h_{l-1}} \mathcal{L}(h_{l-1}, \mathcal{T}).
$$

Since the composite message-passing function is permutation invariant over the neighbors for each node, i.e. $\mathbf{GD}(h_{l-1}, \mathcal{T})$ is permutation invariant over $\mathcal{N}^1[v]$ for any $v \in \mathcal{E}$, we have $\nabla_{h_{l-1}} \mathcal{L}(h_{l-1}, \mathcal{T})$ to be a permutation invariant function. 

Now we show that the corresponding energy function $\mathcal{L}(h_{l-1}, \mathcal{T})$ is permutation invariant over $\mathcal{T}$. Suppose there exist at least one pair of permutations of the triplets $(\mathcal{T}_{\pi_1}, \mathcal{T}_{\pi_2})$
such that
$$\mathcal{L}(h_{l-1}, \mathcal{T}_{\pi_1}) \neq \mathcal{L}(h_{l-1}, \mathcal{T}_{\pi_2}).$$
Then take the derivative w.r.t. $h_{l-1}$, and we get the inequality
$$\nabla_{h_{l-1}}\mathcal{L}(h_{l-1}, \mathcal{T}_{\pi_1}) \neq \nabla_{h_{l-1}}\mathcal{L}(h_{l-1}, \mathcal{T}_{\pi_2}),$$
i.e., there must exist at least one node $v_0$ such that $$\nabla_{h_{l-1}[v_0]}\mathcal{L}(h_{l-1}[v_0], \mathcal{N}^1_{\pi_1'}[v_0]) \neq \nabla_{h_{l-1}[v_0]}\mathcal{L}(h_{l-1}[v_0], \mathcal{N}^1_{\pi_2'}[v_0]),$$
which violates the assumption of MP-GNN. Here $\pi_1'$ and $\pi_2'$ are the subsets of $\pi_1$ and $\pi_2$ including the triplets tailed with $v_0$. Therefore, we get the conclusion that $\mathcal{L}(h_{l-1}, \mathcal{T})$ is permutation invariant over $\mathcal{T}$.

From Deep Sets \cite{zaheer2017deep}, $\mathcal{L}$ can be expressed as 

\begin{equation}
    \mathcal{L}(h_{l-1}, \mathcal{T}) = \rho\left(\sum_{(u, r, v)\in \mathcal{T}} f[(h_{l-1}[u], r, h_{l-1}[v])]\right).
\end{equation}

Since $\mathcal{L}$ is differentiable, take the derivative and we get 

\begin{equation}
    \frac{\partial \mathcal{L}(h_{l-1}, \mathcal{T})}{\partial {h_{l-1}}[v]} = \frac{\partial \rho(\sum_\mathcal{T}f)}{\partial \sum_\mathcal{T}f} \frac{\partial \sum_\mathcal{T}f}{\partial f} \frac{\partial f}{\partial h_{l-1}[v]}.
\end{equation}

If $\rho$ is a nonlinear function, i.e.,
$\rho\left(\sum_{(u, r, v)\in \mathcal{T}} f[(h[u], r, h[v])]\right) \neq \sum_{(u, r, v)\in \mathcal{T}} \rho\left(f[(h[u], r, h[v])]\right)$, the gradient w.r.t. a specific node embedding will use the information over all the triplets, which may violate the message-passing definition. Thus, $\rho$ should be a linear function. Without loss of generality we can assume $\rho$ is an identity mapping, since otherwise we can always find an alternative $f'$ such that $f' = \rho \circ f$. Now we have the energy function in layer $l-1$ of the form

\begin{equation}
    \mathcal{L}(h_{l-1}, \mathcal{T}) = \sum_{(u, r, v)\in \mathcal{T}} f[(h_{l-1}[u], r, h_{l-1}[v])].
\end{equation}

Without loss of generality, from the proof above, the unfolding gradient descent layer for a specific destination node $v$ can be written as

\begin{equation}
    \begin{split}
    h_l[v] &= h_{l-1}[v] - \alpha  \nabla_{h_{l-1}[v]}  \mathcal{L}(h_{l-1}, \mathcal{T}) \\
         &= h_{l-1}[v] - \alpha \left( \sum_{(u, r, v)\in \mathcal{T}} \nabla_{h_{l-1}[v]}  f[(h_{l-1}[u], r, h_{l-1}[v])]\right) \\
         &= h_{l-1}[v] - \alpha \left( \sum_{(u, r)\in \mathcal{N}^1[v]} \nabla_{h_{l-1}[v]}  f[(h_{l-1}[u], r, h_{l-1}[v])]\right). \\
    \end{split}
\end{equation}

\section{Proof of Corollary~1}

Given the energy function $\mathcal{L}(h, \mathcal{T})$ defined in Theorem~1, we take the preconditioner as $\mathbf{D}$ and do the gradient descent over $h[v]$,

\begin{equation}
    \begin{split}
        h_l[v] = \mathbf{PGP}_\mathbf{D}(h_{l-1}, \mathcal{T}) &= h_{l-1}[v] - \alpha \mathbf{D} \nabla_{h_{l-1}[v]}\mathcal{L}(h_{l-1}, \mathcal{T}) \\
        &= h_{l-1}[v] - \alpha \mathbf{D} \nabla_{h_{l-1}[v]}\sum_{(u, r, v) \in \mathcal{T}}f\left((h_{l-1}[u], r, h_{l-1}[v])\right). \\
    \end{split}
\end{equation}

We first show all the functions for $\mathbf{D}$'s rows can be decomposed in the form of Deep Sets \cite{zaheer2017deep} with positive constraints. From the proof of Theorem~2 in \cite{zaheer2017deep}, we have $\sum_{(u, v) \in \mathcal{N}^1[v_i]}\phi(u, r, v_i)$ constituting a unique representation for every set $\mathcal{X} \in 2^{\aleph}$. Then with the positive constraint on $\rho$, we have such $\rho: \mathbb{R} \rightarrow \mathbb{R}^+$ can always be constructed such that $\mathbf{D}_{i\cdot} = \rho\left(\sum_{(u, v) \in \mathcal{N}^1[v_i]}\phi(u, r, v_i)\right)$.

Then we prove the gradient descent step with $\mathbf{D}$ as the preconditioner is a message passing layer. Let $$m[u, r, v] = \nabla_{h_{l-1}[v]} f(h_{l-1}[u], r, h_{l-1}[v])$$ and we have 

\begin{equation}
    h_l[v] = h_{l-1}[v] - \alpha \mathbf{D} \sum_{(u, r) \in \mathcal{N}^1[v]}m[u, r, v],
\end{equation}

where $$a_l[v] = f_A(\{m_l[u, r, v]: (u, r) \in \mathcal{N}^1[v]\}) = \sum_{ (u, r) \in \mathcal{N}^1[v]}m_l[u, r, v]$$ and $$f_U(h_{l-1}[v], a_l[v]) = h_{l-1}[v] - \alpha \mathbf{D} a_l[v].$$

Therefore, we show that the unfolding preconditioned gradient descent is a message-passing layer.

\section{Proof of Theorem~2}

To show its unfolding step is a message-passing layer, we need to prove:

\begin{enumerate}
    \item The update is permutation invariant over triplets in $\mathcal{T}$;
    \item Only the information from 1-hop neighbors are used for updating.
\end{enumerate}

We can concatenate the auxiliary variables $S$ and embedding $h$, and the update step is given by

\begin{equation}
    \begin{split}
        [S'_l, h_l] &= \left[B \cdot \left[S_{l-1}, \Psi(\mathbf{g}_{l-1})\right], h_{l-1} - \alpha \mathbf{D}(\mathbf{g}_{l-1}, S_l) \varphi(\mathbf{g}_{l-1}, S_l)\right] \\
        &= [B^{[,1]} S_{l-1}, h_{l-1}] + [B^{[,2]} \Psi(\mathbf{g}_{l-1}), - \alpha \mathbf{D}(\mathbf{g}_{l-1}, S_l) \varphi(\mathbf{g}_{l-1}, S_l)] \\
        &= [B^{[,1]} S_{l-1}, h_{l-1}] + [B^{[,2]} \Psi(\mathbf{g}_{l-1}), - \alpha \mathbf{D}(\mathbf{g}_{l-1}, S_l)[\varphi_1(S_l) + \varphi_2(S_l)\mathbf{g}_{l-1}]].
    \end{split}
\end{equation}

The last equality is from the assumption that $\varphi(\cdot)$ is a linear function on $\mathbf{g}$, and $\varphi_1, \varphi_2$ are transformations on the hidden states. Since $\psi$ and $\varphi$ are node-wise functions, they maintain the property of permutation invariance. Thus, the update is a gradient descent step with a preconditioner, where the information of hidden states from the last layer is involved. Thus, its unfolding step is a message-passing layer.



\section{Proof of Theroem~3}

\paragraph{From (i) to (ii)}

We make quadratic approximation to $f$ in each layer $l$ at $h_{l-1}[v]$:

\begin{equation}
    \begin{split}
        h_{l}[v] &= \arg \min_{z \in \mathcal{Z}_{\eta}} \sum_{(u, r) \in \mathcal{N}^1[v]} f((h_{l-1}[u], r, z)) + \eta(z) \\
        &= \arg \min_{z \in \mathcal{Z}_{\eta}} \sum_{(u, r) \in \mathcal{N}^1[v]} \left[ f_{h_{l-1}[v]} + \left(\nabla_{h_{l-1}[v]} f_{h_{l-1}[v]}\right)^T(z - h_{l-1}[v]) + \frac{1}{2\alpha}||z - h_{l-1}[v]||^2 \right] + \eta(z) \\
        &= \arg \min_{z \in \mathcal{Z}_{\eta}} \sum_{(u, r) \in \mathcal{N}^1[v]} \left[\frac{1}{2\alpha}||z - (h_{l-1}[v] -\alpha \nabla_{h_{l-1}[v]} f_{h_{l-1}[v]}) ||^2_2 \right] + \eta(z).
    \end{split}
\end{equation}

Here we abbreviate $f((h_{l-1}[u], r, h_{l-1}[v]))$ as $f_{h_{l-1}[v]}$. From the definition of proximal operators, our unfolding layer is 

\begin{equation}
    \begin{split}
        h_l[v] &= \mathrm{prox}_{\eta, \alpha}\left(h_{l-1}[v] - \alpha \sum_{(u, r) \in \mathcal{N}^1[v]} \left[\nabla_{h_{l-1}[v]} f((h_{l-1}[u], r, h_{l-1}[v])) \right] \right). \\
    \end{split}
\end{equation}

Let $$m[u, r, v] = \nabla_{h_{l-1}[v]} f(h_{l-1}[u], r, h_{l-1}[v]),$$ 

and we have 

\begin{equation}
    \begin{split}
        h_l[v] &= \mathrm{prox}_{\eta, \alpha}\left(h_{l-1}[v] - \alpha \sum_{(u, r) \in \mathcal{N}^1[v]} m[u, r, v] \right), \\
    \end{split}
\end{equation}

where 

$$a_l[v] = f_A(\{m_l[u, r, v]: (u, r) \in \mathcal{N}^1[v]\}) = \sum_{ (u, r) \in \mathcal{N}^1[v]}m_l[u, r, v]$$ and $$f_U(h_{l-1}[v], a_l[v]) = \mathrm{prox}_{\eta, \alpha}\left(h_{l-1}[v] - \alpha a_l[v]\right).$$

\paragraph{From (ii) to (i)}

Given a message passing layer from the proximal gradient descent algorithm, denote $g(h[v], \mathcal{T}) = g((h[u], r, h[v]))$, and we have

\begin{equation}
\label{eq: update_prox}
\begin{split}
        h_l &= \mathrm{prox}_{\eta, \alpha}\left( h_{l-1} - \alpha\nabla_{h_{l-1}} g(h_{l-1}, \mathcal{T}) \right) \\
            &= \arg \min_{z \in \mathcal{Z}_{\eta}} \left[\frac{1}{2\alpha} || h_{l-1} - \alpha\nabla_{h_{l-1}} g(h_{l-1}, \mathcal{T}) - z||^2 + \eta(z) \right] \\
            &= \arg \min_{z \in \mathcal{Z}_{\eta}} \left[g(h_{l-1}, \mathcal{T}) + \nabla_{h_{l-1}}g(h_{l-1}, \mathcal{T})^T(z - h_{l-1}) + \frac{1}{2\alpha} || h_{l-1}  - z||^2 + \eta(z) \right] \\
            &= \arg \min_{z \in \mathcal{Z}_{\eta}} \left[g(z, \mathcal{T}) + \eta(z) \right],
\end{split}
\end{equation}

where $g$ is a convex and differentiable function, and $\eta$ is a convex but non-differentiable function. From Eq~\eqref{eq: update_prox}, the energy function is $\mathcal{L}(h_l, \mathcal{T}) = g(h_l, \mathcal{T}) + \eta(h_{l})$. 

From the proof of Theorem~1, if $\mathcal{L}(h_l, \mathcal{T})$ is not permutation invariant over $\mathcal{T}$, $\nabla_{h_{l}[v]}g(h_l[v], \mathcal{T})$ will not be permutation invariant over $\mathcal{N}^1[v]$, which violates the assumption of the message passing function. Thus, $\mathcal{L}(h_l, \mathcal{T})$ should be permutation invariant over $\mathcal{T}$, which is equivalent to the permutation invariance of $g(h_l, \mathcal{T})$ over $\mathcal{T}$. 

With the same route in the proof of Theorem~1, there exists a transformation $f$ such that

\begin{equation}
    g(h_{l}, \mathcal{T}) = \sum_{(u, r, v) \in \mathcal{T}} f[(h_{l}[u], r, h_{l}[v])].
\end{equation}

Thus, with the assumption that $\eta(\cdot)$ is separable, we have the loss for each layer $l$:

\begin{equation}
\begin{split}
    \mathcal{L}(h_{l}, \mathcal{T}) &= \sum_{(u, r, v) \in \mathcal{T}} f[(h_{l}[u], r, h_{l}[v])] + \eta(h_{l}) \\
    &= \sum_{(u, r, v) \in \mathcal{T}} f[(h_{l}[u], r, h_{l}[v])] + \sum_{v \in \mathcal{E}}\eta(h_{l}[v]).
\end{split}
\end{equation}

\section{Examples Derived from Theorem~2}

\subsection{Momentum Accelerated Gradient}
\label{sec: momentum}
Momentum is a method that helps to accelerate gradient descent in the relevant direction and dampens oscillations, and the algorithm in our setting is given by 

\begin{equation}
    \begin{split}
        o_{l} &= \beta o_{l-1} + \alpha \nabla_{h_{l-1}}\mathcal{L}(h_{l-1}, \mathcal{T}), \\
        h_l &= h_{l-1} - o_l.
    \end{split}
\end{equation}

To show that we can use momentum to form a message-passing layer, we concatenate $o$ and $h$ by columns, i.e., we have $y = [o, h] \in \mathbb{R}^{|\mathcal{E}|\times 2d}$. Then we have

\begin{equation}
    \begin{split}
        [o_l, h_l] &= [\beta o_{l-1} + \alpha \nabla_{h_{l-1}}\mathcal{L}(h_{l-1}, \mathcal{T}), h_{l-1} - o_l] \\
        &= [0, h_{l-1}] + \left[\beta o_{l-1} +  \alpha \nabla_{h_{l-1}}\mathcal{L}(h_{l-1}, \mathcal{T})\right][1, -1]. \\
    \end{split}
\end{equation}

For a specific node $v$, the update function is 

\begin{equation}
    \begin{split}
        [o_l[v], h_l[v]] &= [0, h_{l-1}[v]] + \left[\beta o_{l-1}[v] +  \alpha \sum_{(u, r, v) \in \mathcal{T}} \nabla_{h_{l-1}[v]} f\left((h_{l-1}[u], r, h_{l-1}[v])\right)\right][1, -1].
    \end{split}
\end{equation}

Let $$m_{l-1}[v] = \nabla_{h_{l-1}[v]} f\left((h_{l-1}[u], r, h_{l-1}[v])\right),$$

$$
a_{l-1}[v] = \beta o_{l-1}[v] + \alpha  \sum_{(u, r, v) \in \mathcal{T}} m_{l-1}[v],
$$

and the update function is

\begin{equation}
    \begin{split}
        [o_l[v], h_l[v]] &= [0, h_{l-1}[v]] + [1, -1]a_{l-1}[v].
    \end{split}
\end{equation}

Thus, in this case the velocity $o_{l}$ is the output of the aggregation function from the previous layer $l-1$. Since the aggregation function is permutation invariant and the unfolding step only uses 1-hop neighbors of every node, we have $\mathcal{L}$'s unfolding gradient descent step with momentum is a message-passing layer.

\subsection{Adagrad}
\label{sec: adagrad}

Adagrad uses a different learning rate for every parameter. We denote $$\mathbf{g}_l[v] = \nabla_{h_{l}[v]}\mathcal{L}(h_l[v], \mathcal{T}) = \nabla_{h_l[v]} \sum_{(u, r, v) \in \mathcal{T}} f\left((h_l[u], r, h_l[v])\right).$$

The update rule of Adagrad is 

\begin{equation}
    \begin{split}
        s_l[v] &= s_{l-1}[v] + \mathbf{g}_{l-1}^2[v], \\
        h_{l}[v] &= h_{l-1}[v] - \frac{\alpha}{\sqrt{s_{l}[v]+\epsilon}}\mathbf{g}_{l-1}[v].
    \end{split}
\end{equation}

Use the same strategy in Sec~\ref{sec: momentum}, we have 

\begin{equation}
    \begin{split}
        [s_l[v], h_l[v]] &= \left[s_{l-1}[v] + \mathbf{g}_{l-1}^2[v], h_{l-1}[v] - \frac{\alpha}{\sqrt{s_{l-1}[v] + \mathbf{g}_{l-1}^2[v]+\epsilon}}\mathbf{g}_{l-1}[v] \right].
    \end{split}
\end{equation}

Here, let 

$$
m_{l-1}[v] = \nabla_{h_{l-1}[v]} f\left((h_{l-1}[u], r, h_{l-1}[v])\right)
$$

and

$$
a_{l-1}[v] = \mathbf{g}_{l-1}[v] = \sum_{(u, r) \in \mathcal{N}^1[v]} m_{l-1}[v],
$$

and we denote 

$$
\mathbf{D}_{l-1}[v] = \sqrt{s_{l-1}[v] + a^2_{l-1}[v]+ \epsilon}.
$$

Then the update function is 

\begin{equation}
    \begin{split}
        [s_l[v], h_l[v]] &= [s_{l-1}[v], h_{l-1}[v]]  +  \left[a_{l-1}[v],  - \frac{\alpha}{\mathbf{D}_{l-1}[v]} \right] a_{l-1}[v].
    \end{split}
\end{equation}

\subsection{RMSprop}

The update rule of RMSprop is 

\begin{equation}
    \begin{split}
        s_l &= \beta s_{l-1} + (1-\beta)\mathbf{g}_{l-1}^2, \\
        h_{l} &= h_{l-1} - \frac{\alpha}{\sqrt{s_l+\epsilon}} \mathbf{g}_{l-1}.
    \end{split}
\end{equation}

Thus, we have

\begin{equation}
    \begin{split}
        [s_l[v], h_l[v]] &= \left[\beta s_{l-1}[v] + (1-\beta)\mathbf{g}_{l-1}^2[v], h_{l-1}[v] - \frac{\alpha}{\sqrt{\beta s_{l-1}[v] + (1-\beta)\mathbf{g}_{l-1}^2[v]+\epsilon}}\mathbf{g}_{l-1}[v] \right]\\
        &= \left[\beta s_{l-1}[v] , h_{l-1}[v]  \right] + \left[(1-\beta)\mathbf{g}_{l-1}^2[v], - \frac{\alpha}{\sqrt{\beta s_{l-1}[v] + (1-\beta)\mathbf{g}_{l-1}^2[v]+\epsilon}}\mathbf{g}_{l-1}[v]\right]\\
        &= \left[\beta s_{l-1}[v] , h_{l-1}[v]  \right] + \left[(1-\beta)a_{l-1}[v], - \frac{\alpha}{\mathbf{D}_{l-1}[v]}\right] a_{l-1}[v],\\
    \end{split}
\end{equation}

where

$$
a_{l-1}[v] = \mathbf{g}_{l-1}[v] = \sum_{(u, r) \in \mathcal{N}^1[v]} m_{l-1}[v] = \sum_{(u, r) \in \mathcal{N}^1[v]}\nabla_{h_{l-1}[v]} f\left((h_{l-1}[u], r, h_{l-1}[v])\right)
$$

and

$$
\mathbf{D}_{l-1}[v] = \sqrt{\beta s_{l-1}[v] + (1-\beta)\mathbf{g}_{l-1}^2[v]+\epsilon}.
$$



\subsection{Adam}
The update rule of Adam is 

\begin{equation}
    \begin{split}
        o_l &= \beta_1 o_{l-1} - (1-\beta_1) \mathbf{g}_{l-1}, \\
        s_l &= \beta_2 s_{l-1} + (1-\beta_2) \mathbf{g}_{l-1}^2, \\
        h_{l} &= h_{l-1} - \alpha \frac{\hat{o}_{l}}{\sqrt{\hat{s}_{l}} + \epsilon},
    \end{split}
\end{equation}

where $\hat{o}_l = \frac{o_l}{1 - \beta_1^l}$ and $\hat{s}_l = \frac{s_l}{1 - \beta_2^l}$.

Here we concatenate $o, s, h$ and get

\begin{equation}
    \begin{split}
        [o_{l}, s_{l}, h_{l}] =& \left[ \beta_1 o_{l-1} - (1-\beta_1) \mathbf{g}_{l-1}, \beta_2 s_{l-1} + (1-\beta_2) \mathbf{g}_{l-1}^2, h_{l-1} - \alpha \frac{\hat{o}_{l}}{\sqrt{\hat{s}_{l}} + \epsilon} \right] \\
        =& [\beta_1 o_{l-1}, \beta_2 s_{l-1}, h_{l-1}] - \left[ (1-\beta_1) \mathbf{g}_{l-1}, -(1-\beta_2) \mathbf{g}_{l-1}^2, \alpha \frac{\frac{\beta_1 o_{l-1} - (1-\beta_1) \mathbf{g}_{l-1}}{1 - \beta_1^l}}{\sqrt{\frac{\beta_2 s_{l-1} + (1-\beta_2) \mathbf{g}_{l-1}^2}{1 - \beta_2^l}} + \epsilon} \right].
    \end{split}
\end{equation}

We write 

\begin{equation}
    \frac{\frac{\beta_1 o_{l-1} - (1-\beta_1) \mathbf{g}_{l-1}}{1 - \beta_1^l}}{\sqrt{\frac{\beta_2 s_{l-1} + (1-\beta_2) \mathbf{g}_{l-1}^2}{1 - \beta_2^l}} + \epsilon} = \frac{\beta_1 o_{l-1}}{\mathbf{D}_{l-1}} - \frac{1-\beta_1}{\mathbf{D}_{l-1}}a_{l-1},
\end{equation}

where $$
\mathbf{D}_{l-1} = (1 - \beta_1^l) \left(\sqrt{\frac{\beta_2 s_{l-1} + (1-\beta_2) a_{l-1}^2}{1 - \beta_2^l}} + \epsilon\right)
$$

and

$$
a_{l-1}[v] = \mathbf{g}_{l-1}[v] = \sum_{(u, r) \in \mathcal{N}^1[v]} m_{l-1}[v] = \sum_{(u, r) \in \mathcal{N}^1[v]}\nabla_{h_{l-1}[v]} f\left((h_{l-1}[u], r, h_{l-1}[v])\right).
$$

\bibliographystyle{plain}
\bibliography{ref}